\pdfoutput=1
\documentclass{CUP-JNL-EDS}
\usepackage{latexsym}
\usepackage{graphicx}
\usepackage{multicol,multirow}
\usepackage{amsmath,amssymb,amsfonts}
\usepackage{mathrsfs}
\usepackage{amsthm}
\usepackage{rotating}
\usepackage{appendix}
\usepackage[authoryear]{natbib}
\usepackage{ifpdf}
\usepackage[T1]{fontenc}
\usepackage{times}
\usepackage{newtxmath}
\usepackage{textcomp}
\usepackage{xcolor}
\usepackage{hyperref}
\usepackage{lipsum}
\newcommand{\attention}[1]{{\color{blue}{[#1]}}}
\newcommand{\remove}[1]{ }

\usepackage{makecell}
\PassOptionsToPackage{table,xcdraw}{xcolor}
\usepackage[table]{xcolor}
\articletype{UNDER REVIEW - AS SURVEY ARTICLE}
\jname{Environmental Data Science}
\jyear{202x}
\jvol{x}
\jdoi{10.1017/eds.xxxx.xx}

\begin{document}
\begin{Frontmatter}

\title[Article Title]
{Tools for Extracting Spatio-Temporal Patterns in 
Meteorological Image Sequences: From Feature Engineering to Attention-Based Neural Networks}

\author*[1]{Akansha Singh Bansal}\email{akanshasingh.bansal@colostate.edu}\orcid{0000-0002-9637-7800}
\author[1]{Yoonjin Lee}\orcid{0000-0002-2092-3078}
\author[1]{Kyle Hilburn}\orcid{0000-0002-2078-9884}
\author[1,2]{Imme Ebert-Uphoff}\orcid{0000-0001-6470-1947}
\address[1]{\orgdiv{Cooperative Institute for Research in the Atmosphere}, \orgname{Colorado State University}, \orgaddress{\city{Fort Collins}, \state{Colorado},  \country{USA}}}
\address[2]{\orgdiv{Electrical and Computer Engineering}, \orgname{Colorado State University}, \orgaddress{\city{Fort Collins}, \state{Colorado},  \country{USA}}}

\received{20 October 2022}

\authormark{Bansal et al.}
\keywords{Meteorological Data, Spatio-Temporal Patterns, Image Sequence, Satellite Imagery, Remote Sensing, Machine Learning, Artifical Intelligence, Attention}

\abstract{
Atmospheric processes involve both space and time.  This is why human analysis of atmospheric imagery can often extract more information from animated loops of image sequences than from individual images.  Automating such an analysis requires the ability to identify spatio-temporal patterns in image sequences which is a very challenging task, because of the endless possibilities of patterns in both space and time. In this paper we review different concepts and techniques that are useful to extract spatio-temporal context specifically for meteorological applications.  In this survey we first motivate the need for these approaches in meteorology using two applications, solar forecasting and detecting convection from satellite imagery.  Then we provide an overview of many different concepts and techniques that are helpful for the interpretation of meteorological image sequences, such as  (1) feature engineering methods to strengthen the desired signal in the input, using meteorological knowledge, classic image processing, harmonic analysis and topological data analysis; (2) explain how different convolution filters (2D/3D/LSTM-convolution) can be utilized strategically in convolutional neural network architectures to find patterns in both space and time; (3) discuss the powerful new concept of “attention” in neural networks and the powerful abilities it brings to the interpretation of image sequences; (4) briefly survey strategies from unsupervised, self-supervised and transfer learning to reduce the need for large labeled datasets.  We hope that presenting an overview of these tools – many of which are underutilized - will help accelerate progress in this area. 
}

\policy{We hope that this survey will help scientists develop algorithms that can detect and model complex spatio-temporal patterns inherent in  meteorological image sequences.  This would allow machine learning algorithms to (1) identify more complex phenomena primarily visible in image loops (such as convection), and (2) provide predictions across longer time spans, than currently possible.}

\end{Frontmatter}

\section{Introduction}
\label{sec:introduction}
\begin{figure*}
\begin{center}
\centerline{\includegraphics[width=1\linewidth]{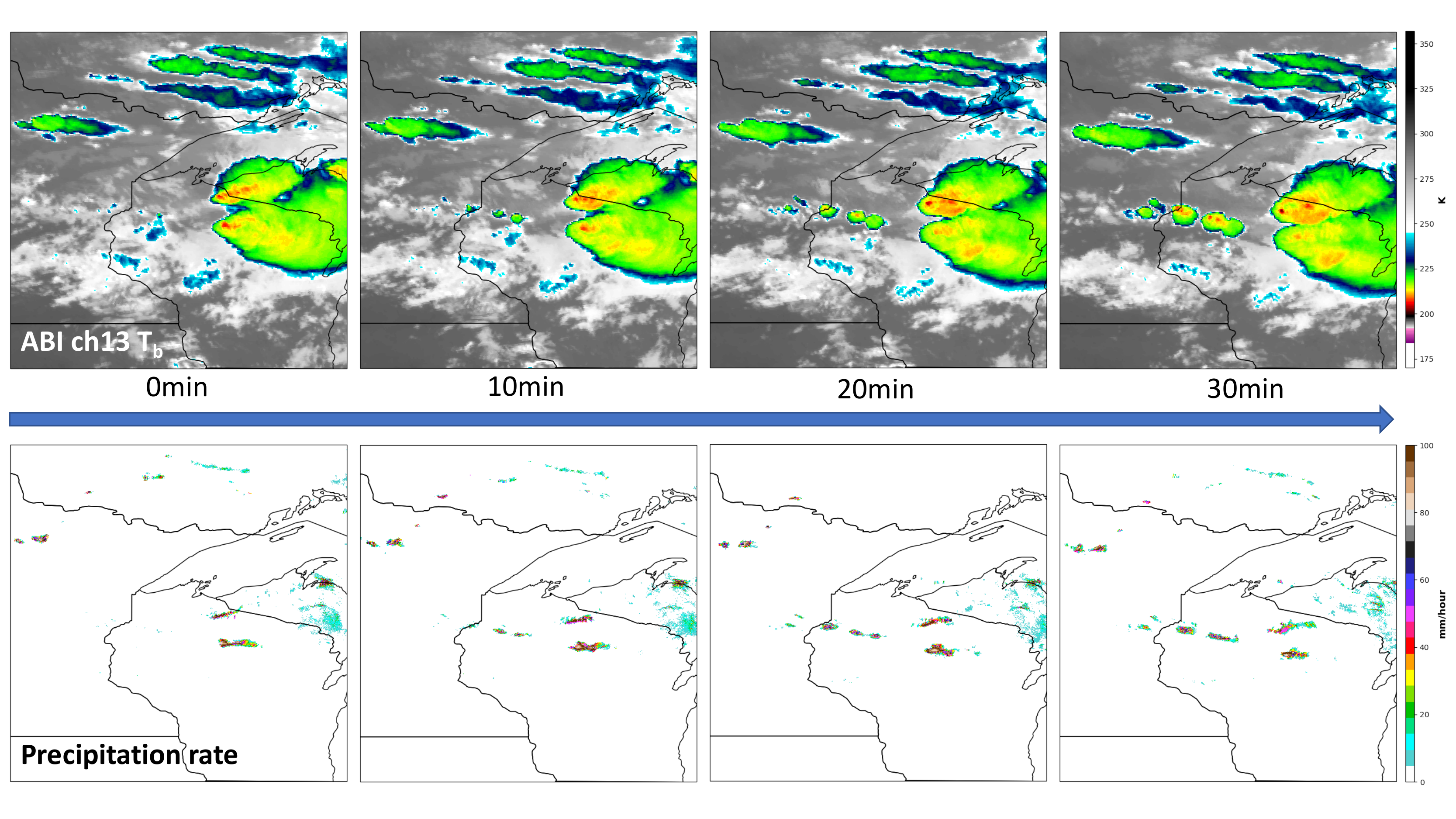}}
\vspace{-0.3cm}
\caption{Sample meteorological image sequence that provided important context. Top row: Geostationary Operational Environmental Satellites (GOES)-16 Advanced Baseline Imager (ABI) Channel 13 brightness temperature.  Bottom row: corresponding Multi-Radar/Multi-Sensor System (MRMS) precipitation rate.  This sequence shows the evolution of convective clouds, which is important to predict the future behavior of the developing convection}
\label{fig:abi_mrms}
\end{center}
\vspace{-0.6cm}
\end{figure*}

Meteorological imagery is used routinely to extract, and often to predict, certain states of the earth system, e.g., for applications ranging from weather forecasting to wildfire detection. Such meteorological imagery can come from numerical models (e.g., GCMs) or remote sensing (e.g., satellites or radar).  Spatio means space and temporal means time so put together, spatio-temporal means information within space {\it and} time.  The value of spatio-temporal information content in meteorological imagery is why human analysts can extract more information from animated loops of image sequences than from an individual image. For example, Figure \ref{fig:abi_mrms} shows a meteorological image sequence that captures the  evolution of convective clouds. We can observe cloud top cooling in GOES-16 ABI image sequences followed by increase in precipitation rate. Knowledge of this evolution is helpful to identify convection at the last time step and to predict the further development of convection in the next time step.

This information is especially valuable when different phenomena have similar radiometric signatures, such as bright clouds moving through the scene versus bright snow on the ground; or when phenomena are present at different vertical levels, such as low and high clouds moving at different speeds and directions. This article provides a survey of data science approaches that can be used to extract signals that span both time and space from such imagery.

When extracting information from image sequences a key question is what kind of meteorological context is required to perform the desired task, such as:
\begin{itemize}
\item
   {\bf Spatial context:} Looking for spatial patterns in the imagery allows one to interpret values at each pixel knowing that it is part of a larger meteorological pattern, e.g., part of a tropical cyclone; 
\item
   {\bf Temporal context:} Utilizing several time steps can be useful to gain information on the temporal evolution of a phenomenon.
\item 
   {\bf Multi-variable context (spectral channels or atmospheric variables):} 
   Using several satellite channels (if the imagery comes from satellites) or different atmospheric variables (if the imagery comes from models) is often helpful since different variables capture different aspects of the atmospheric (or environmental) state.  
\end{itemize}

Many meteorological applications require spatial, temporal and multi-variable context.  For example, a human can identify convection more easily from sequences of satellite imagery, rather than from a single image. 
Likewise, 
it has been shown that neural networks can also perform this task better when given temporal {\it sequences} of imagery, as shown by  \citet{yoonjin4}.  Multi-spectral context is also relevant in this application, as both visible and infrared channels provide important information. 
Similarly, forecasting solar radiation also benefits from providing a sequence of temporal images to a neural network architecture, as shown by \cite{bansal2021moment, bansal2021self}.  These two applications are discussed in more detail in Section \ref{sample_applications_sec}.

\label{sec:symbols}

\begin{figure}
    \centering
    \includegraphics[width = 0.8\textwidth]{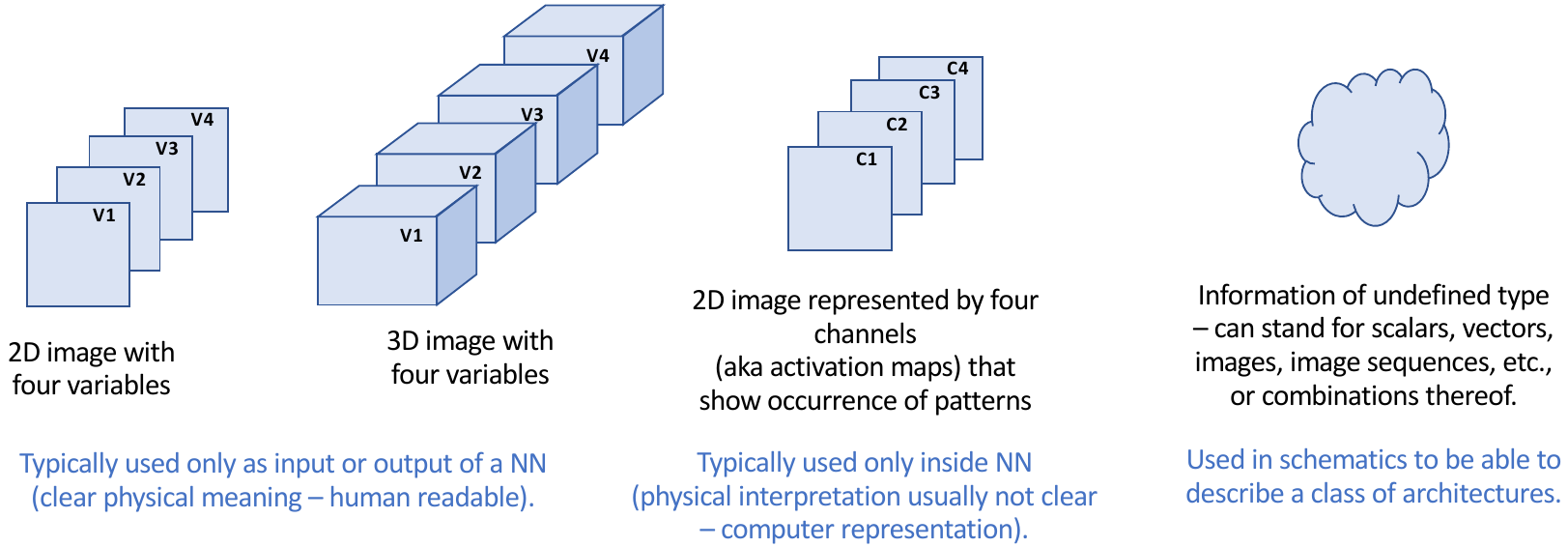}
    
    \hspace*{1.0cm} (a) Representation of 2D and 3D images \hspace*{3.0cm} (b) Undefined type
    \\[0.5cm]

    \includegraphics[width = 0.20 \textwidth]{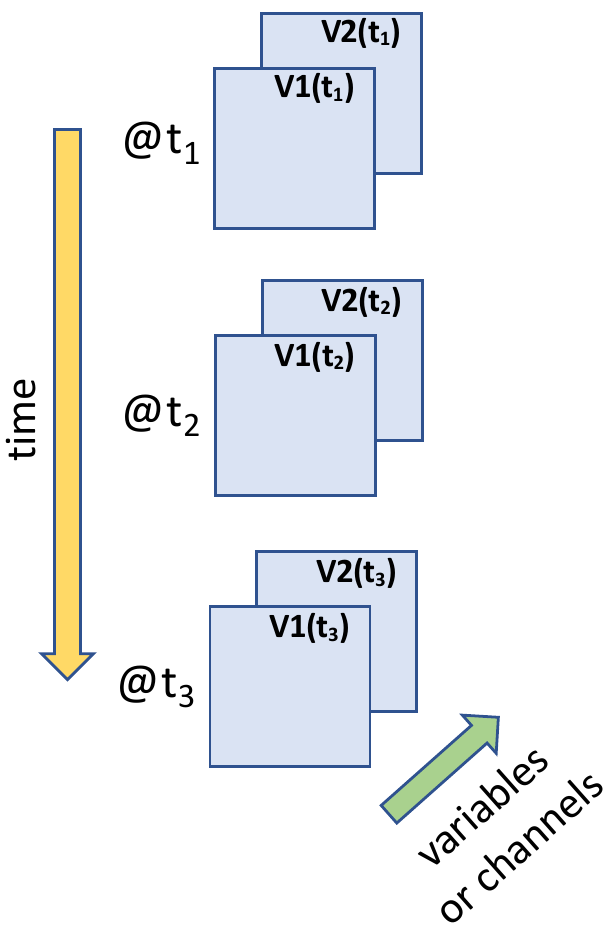}
    \hspace*{2.5cm}
    \includegraphics[width = 0.21 \textwidth]{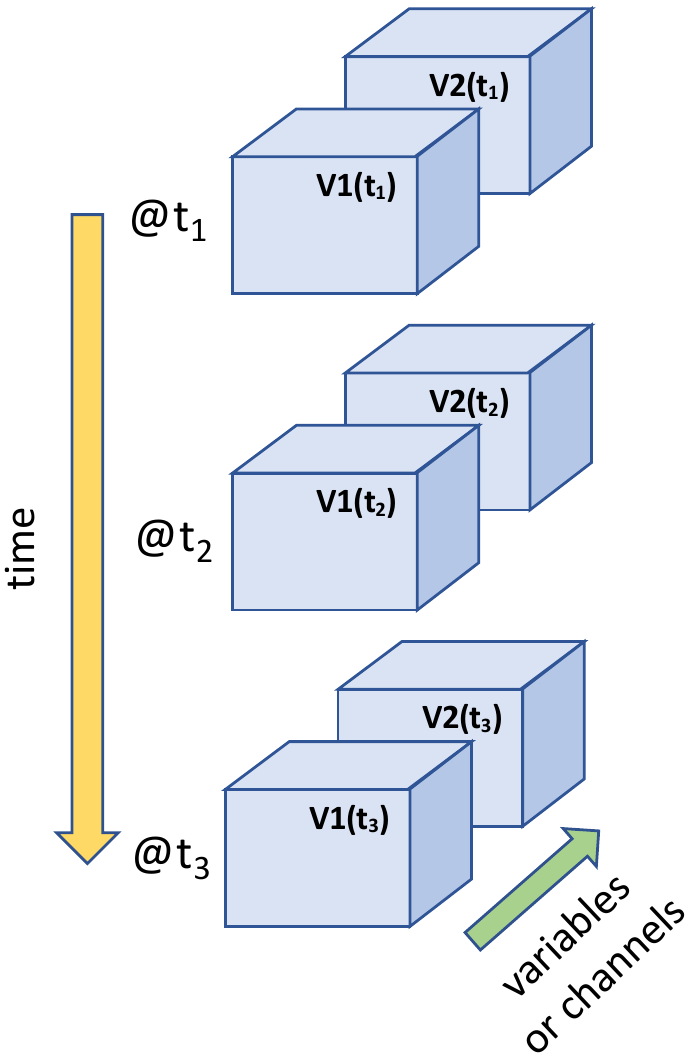}
    \\
    (c) Sequence of 2D images \hspace*{2.0cm} (d) Sequence of 3D images
    \caption{Visual representation of a single image (a), undefined type (b), and of image sequences (c,d) used in this article.  For image sequences, note that time is {\bf always} represented vertically (direction of yellow arrow) and variables/channels are shown as stacks of individual 2D or 3D elements (direction of green arrow).  The specific number of variables or channels  shown here (four) is only used for illustration, in particular to demonstrate that the number of channels does not have to be three, as is typically assumed in computer vision applications  since they tend to deal with sequences of RGB images}
    \label{fig:image_symbols}
    \label{fig:image_sequence}
\end{figure}

Figure \ref{fig:image_symbols} illustrates how we represent images and image sequences throughout this article.
Note in particular the directions of time (downward, yellow arrow) versus variable/channel dimension (stacked, green arrow) used for image sequences, as shown in Figure \ref{fig:image_sequence}(c,d).

\begin{figure}
    \centering
    \includegraphics[width = 1 \textwidth]{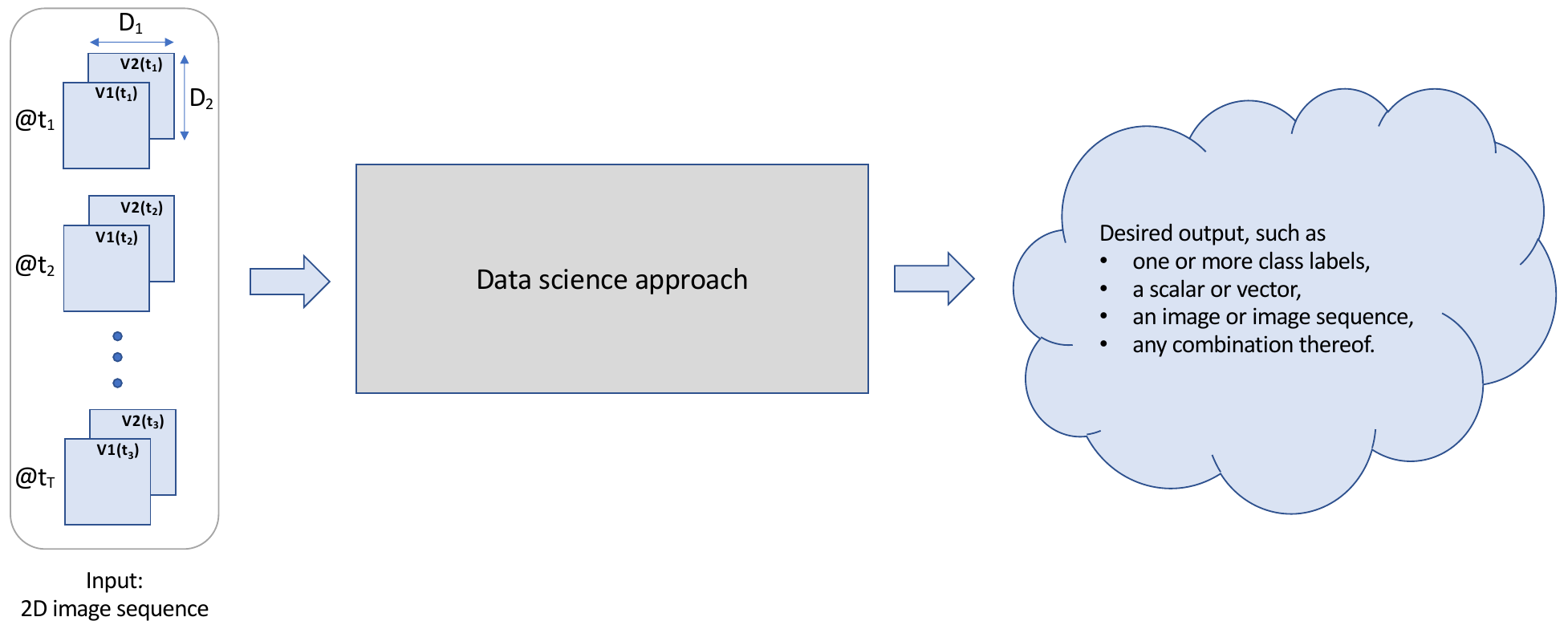} 
    \caption{Problem definition for extracting spatio-temporal information from a meteorological image sequence. The dimension names of the input image sequence shown here are used throughout this document: $D_i$ for spatial dimensions, $T$ for number of time steps, and $V$ (here $V=2$) for the number of variables}
    \label{fig:problem_def}
\end{figure}

Figure\ \ref{fig:problem_def} provides a high level overview of the problem we consider.  We are given a meteorological image sequence with three types of dimensions, namely, two or more spatial dimensions ($D_i$), a temporal dimension (T), and a multi-variable dimension (V).  The task we want to achieve is to use a combination of feature engineering and machine learning to extract those spatio-temporal signals from the image sequence relevant to estimate the desired output.  The desired output may be a class label (e.g., precipitation type), a scalar (e.g., estimated wind speed), an image (e.g., simulated radar \citep{ebert2020evaluation}), or any other desired type of information. 

The key challenge to design a good method to detect all relevant spatio-temporal patterns lies in the high dimensionality of the pattern space, namely of the set of all {\it combinations} of real values extending over the entire space of spatial, temporal and variable dimensions.  For a 2D image sequence that means searching for all possible patterns in a multi-dimensional array of dimension $(D1, D2, T, V)$.  Searching the entire space is not feasible, and thus one must find simplifications that allow us to reduce the search space while still being able to capture all physically relevant patterns of the task at hand. 

Which method is best suited to extract relevant spatio-temporal patterns depends on various properties of the considered task, including: 
\begin{enumerate}
\item 
    {\bf How complex are the patterns to be detected?}\\
    Simple pattern example: Area of high contrast that persists in single location over two time steps.\\
    Complex pattern example: Cloud bubbling pattern indicating convection (see Section \ref{sec:convection}).  
\item 
   {\bf Is prior knowledge available regarding the patterns to be detected?}\\
   If so we can design simplified methods that only focus on a prescribed family of patterns.
\item 
    {\bf How far does each pattern extend in space and time?}\\
    Example: Is it sufficient  to consider values in close spatial and temporal proximity of each other?  If so then dimensionality of the pattern search space is greatly reduced. 
\item
    {\bf Can pattern detection be decoupled in space and time?}\\
    Example: Is it sufficient for the task to look for patterns in each image individually, then track their occurrence over time? 
\item   
   {\bf How many labeled samples are available for training?}
\item
   {\bf How important is transparency to the end user, i.e.\ an understanding of the strategies used by the ML model to perform its task?}
\item
   {\bf Are any models available that have been trained on similar tasks and that could provide a starting point for this task?}
\end{enumerate}

\begin{figure}
\begin{center}    
   
   Tools for extracting spatio-temporal patterns in image sequences \hspace*{0,3cm}
    \includegraphics[width=1\linewidth]{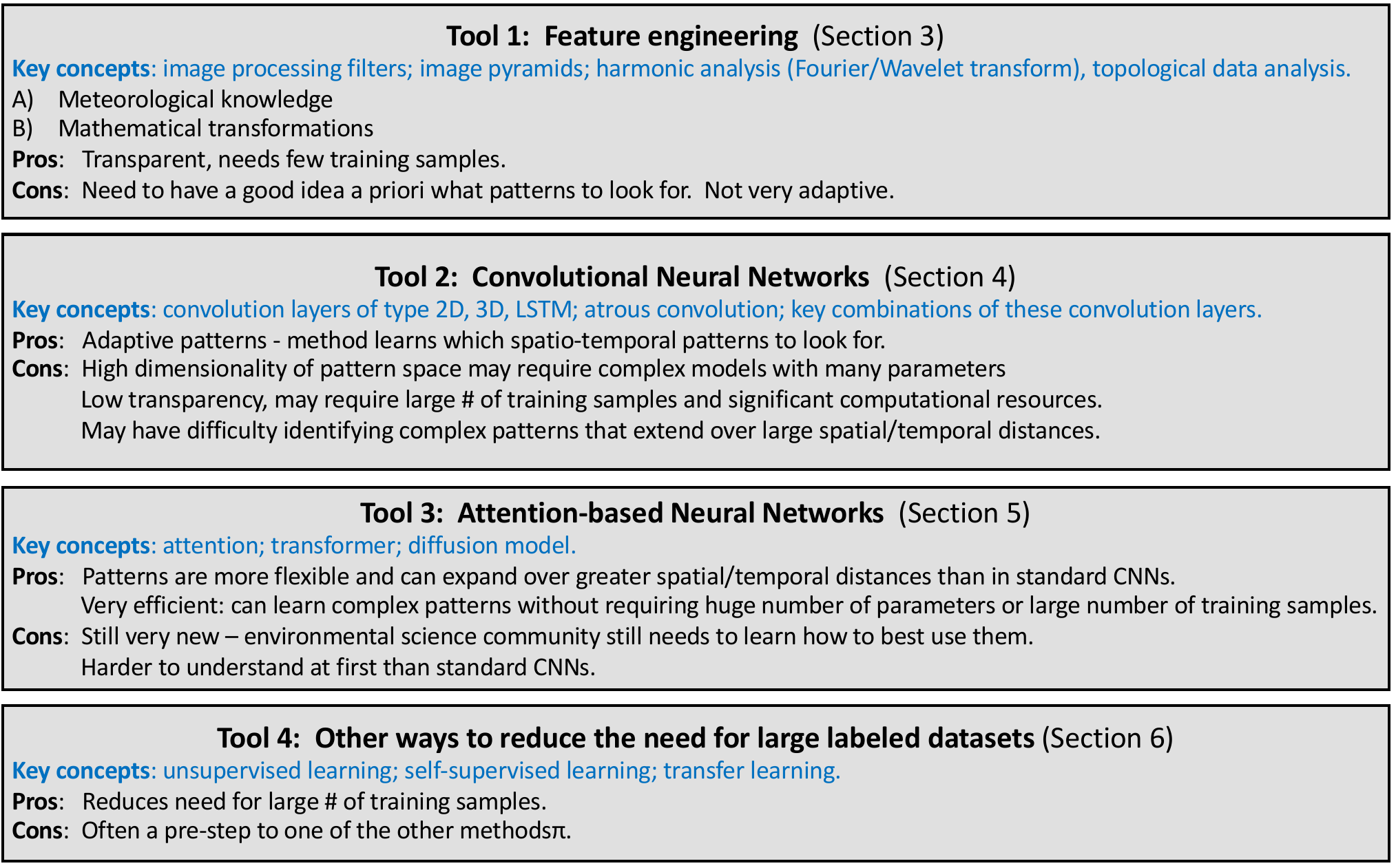}
    \caption{Overview of four types of tools discussed here, including key concepts, their pros and cons, and the sections where they are introduced}
    \label{fig:overview_of_concepts}
\end{center}
\end{figure}

The remainder of this article is organized as follows. 
Section \ref{sample_applications_sec} provides some practical examples from our own research that demonstrate the need for extracting spatio-temporal context from meteorological imagery. 
In Sections \ref{sec:feature_engineering} to \ref{sec:training} we discuss four groups of approaches that each address different needs and have their own pros and cons, as outlined in Figure \ref{fig:overview_of_concepts}.
We structure the discussion of approaches for incorporating spatial context based on how much expert knowledge they use, starting with the highest level of expert knowledge used in the feature engineering section, Section \ref{sec:feature_engineering}. Sections \ref{sec:NN_simple} and \ref{sec:attention_new} discuss increasingly complex machine learning methods from traditional convolutional neural networks (Section \ref{sec:NN_simple}) to advanced neural networks that use attention (Section \ref{sec:attention_new}).  Section \ref{sec:training_strategies} very briefly discusses different strategies that can help reduce the need for a large number of labeled samples for training, such as unsupervised learning and transfer learning. Section \ref{sec:conclusion} presents conclusions.

\section{Sample Applications illustrating importance of spatio-temporal context}
\label{sample_applications_sec}
\label{sec:applications}

In this section we discuss two sample applications that demonstrate the need to extract spatio-temporal context from image sequences.  The two applications both use geostationary satellite image sequences as input, namely (1) forecasting solar radiation, and (2) detecting convection.  Satellite applications are chosen as examples here simply because that is a primary research topic of our group.

\subsection{Nowcasting solar radiation}

Solar power is the cheapest form of electricity in history but its contribution in the electric grid remains low. This is mainly because of solar's variability that changes with the amount of downwelling radiation that reaches the Earth's surface. Effectively predicting the amount of irradiance is akin to forecasting the amount of solar energy produced by the solar panels because of the strong correlation beween the two
~\citep{raza2016recent}. Many techniques have been proposed in the past to address this problem including Numerical Weather Predictions (NWP) algorithms \citep{10.1115/1.4042972,8245549,8447751,MATHIESEN2011967}, that mostly leverage physics-based modeling. These are often used for solar irradiance forecasting, and  
are most appropriate for forecast horizons on the scale of hours to days, and not near-term forecasts on the scale of minutes to an hour \citep{hao2019novel,wang2019review}. 

\begin{figure*}
\begin{center}
\centerline{\includegraphics[width=1\linewidth]{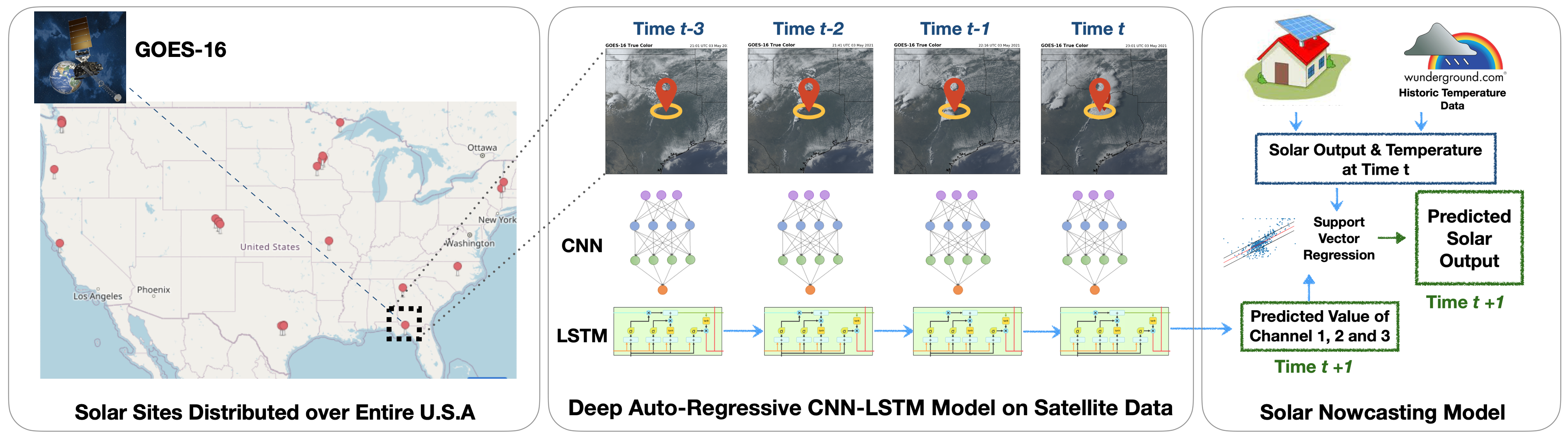}}
\vspace{-0.3cm}
\caption{Overview of the two-step spatio-temporal CNN-LSTM modeling pipeline. The authors collect multispectral satellite data from 25 U.S. sites and use sequences of observations at each site, specifically the first 3 spectral channels (visualized in the middle), as input to a CNN-LSTM model, which they train to predict the observation at the center (red pin) for the next time step ($t+1$). An auto-regressive SVR then takes the predicted channels and previous solar output, as well as previous temperature, to predict the solar output at next step, see \cite{bansal2021moment} and Section \ref{sec:CNN-LSTM} for more explanation}
\label{fig:design}
\label{fig:Akansha_original_image}
\end{center}
\vspace{-0.6cm}
\end{figure*}

The other way of tackling this problem is with machine learning approaches that have the potential to implicitly model local changes directly from observational data \citep{wang2019review,rolnick2019tackling} like analyzing images from ground-based sky cameras \citep{zhang2018deep,zhao20193d,siddiqui2019deep,paletta2020convolutional} and estimating cloud motion vectors \citep{lorenz2004short, lorenz2012prediction, cros2014} from satellite images. However, installing sky cameras requires additional infrastructure making the approach less scalable. Instead, ML approaches that more directly model solar irradiance tend to perform better \citep{lago2018short,dsr} in comparison to the other approaches. \cite{bansal2021moment} propose an approach for solar nowcasting by forecasting solar irradiance values from multispectral visible bands from GOES satellite data using a combined CNN-LSTM model (see Figure\ \ref{fig:design} and explanation in Section \ref{sec:CNN-LSTM}). This model forecast changes in satellite's spatial features over time. Their self-supervised approach directly uses abundant satellite data for modeling to capture its spatio-temporal properties that effectively maps the amount of solar irradiance received on the Earth's surface. They further combine these solar irradiance forecasts with a model that predicts a site's solar output from solar irradiance.

The performance gain from using such spatio-temporal models clearly demonstrates the benefits of extracting spatio-temporal patterns from image sequences for this application.

\subsection{Detecting convection from geostationary satellite}
\label{sec:convection}
Convection is a rapidly developing feature within a few hours, and being able to observe the correct location of convection is critical in short-term forecasts. Radar reflectivity from a ground-based radar is a good indicator of precipitation intensity, hence convection, and it is the main observation used to detect convection. Despite its high accuracy and high temporal resolution data which is suitable for short-term forecasts, it has its own limitation of less coverage over mountainous regions and ocean. Outside of ground-based radar coverage, a geostationary satellite is the only observation available for convection detection. It is challenging to use geostationary satellites as its visible or infrared data only provide cloud top information, but its high temporal resolutions allow us to better observe the development of convective clouds. With the help of machine learning techniques it has become easier to extract spatial and temporal features of convective clouds. 
Features of convective clouds that can be observed by visible and infrared sensors on geostationary satellites are: high reflectance, low brightness temperature, bubbling cloud top, and decrease in brightness temperature. We can somewhat detect convection by looking at high reflectance and low brightness temperature from a static image, but the bubbling cloud top signal stands out much more in temporal image sequences. Similarly, decrease in brightness temperature is a feature that can only be observed by the temporal image sequences. \cite{yoonjin4} explores use of temporal image sequences from GOES-16 to detect convection, and the results are validated against one of the ground-based radar products called Multi-Radar/Multi-Sensor System (MRMS). It shows that using image sequences provided slightly better results than using a static image. 

\section{Feature Engineering - Constructing strong signals for ML methods to use}
\label{sec:approaches}
\label{sec:feature_engineering}
\label{sec:feature_engineering}

The meteorological community often uses the term {\it (statistical) predictors} to refer to quantities that are deemed important for the prediction of quantities to be estimated (predictands).  In the machine learning literature the predictors are known as {\it features}.  We will use the two terms predictor and feature interchangeably throughout this article.  {\it Feature selection} refers to the process of selecting key predictors from a large set of possible predictors, e.g., choosing which satellite channels contain the most information. 
{\it Feature engineering} goes further by creating new features from the considered set of predictors. For example, teleconnection indices, such as the North Atlantic Oscillation (NAO), West Pacific Oscillation (WPO), East Pacific Oscillation (EPO) \citep{barnston1987}, Pacific/North American Pattern (PNA) \citep{wallace1981}, and Arctic Oscillation (AO) \citep{thompson1998} are scalar features extracted from atmospheric fields, developed over decades to capture key information about the atmospheric state in a minimal representation.  A simpler example is the split window difference of satellite channels, i.e., the difference between two different channels, which experts know contains specific information relevant to a problem, such as dust detection \citep{miller2019} or estimating low-level moisture \citep{dostalek2021}.
More generally, feature engineering seeks to extract key information and make it easily accessible for subsequent models, such as statistical or machine learning models.  Engineered features may achieve dimension reduction - the teleconnection indices are an extreme example of this - or it may retain the original dimension and instead amplify a key signal in the data that makes subsequent analysis more robust and successful.

\subsection{Meteorologically motivated image features (most transparent)}

Meteorologists have developed a rich set of statistical predictors for any meteorological application one can think of.  Over decades such predictors were carefully designed using expert intuition, and refined through testing to yield the best predictions possible when feeding them as inputs into statistical or, more recently, machine learning methods. This includes predictors obtained from imagery. 

For example, in the area of tropical cyclones (TCs), the statistical hurricane intensity prediction scheme (SHIPS; \citet{demaria1994statistical}) defines a set of predictors that are deemed useful to predict properties of tropical cyclones, such as future TC intensity and location.  Newer versions of the SHIPS data base \citep{cirapage}
include scalar predictors obtained from satellite imagery, such as the minimum, maximum, average, or standard deviation of the brightness temperature within certain radii from the storm center; fraction of area with brightness temperatures colder than particular thresholds; or principal component analysis \citep{knaff2016}.

The Warning Decision Support System - Integrated Information, aka {\it WDSS-II}, developed by the National Severe Storm Laboratory (NSSL) is a real time system used by the National Weather Service to analyze weather data \citep{lakshmanan2007}. This system uses morphological image processing techniques to extract storm cells from images \citep{lakshmanan2009}, track those cells in time \citep{lakshmanan2010}, and compute statistics and time trends of storm attributes for the cells \citep{lakshmanansmith2009}.

The advantages of using meteorologically motivated features are that they may provide dimensionality reduction, enhance the strength of key signals, and provide models that are physically interpretable.  We believe that feature engineering methods are sometimes under-utilized when advanced machine learning methods are employed. Feature engineering should be kept in mind as a powerful means to infuse physical knowledge into machine learning, and thus to simplify ML models and improve their transparency and robustness. 

\subsection{Mathematically motivated image features (moderately transparent)}

This section briefly discusses three mathematical frameworks that can help to extract spatial features of images in the context of neural networks.  We feel that all of these methods are currently under-utilized in the environmental science and should receive more attention. This is by no means a complete list, we are certain that there are other important mathematical tools we have not thought of.  The three frameworks discussed below only cover those our group has tried for environmental applications. 

\subsubsection{\bf Classic image processing tools and image pyramids}
\label{sec:classic_image_processing_tools}

Traditional satellite retrieval techniques tend to either treat individual satellite pixels as being independent or they make use of simple spatial information, such as the standard deviation in a neighborhood \citep{grecu2001}. The information content in spatial patterns is an important factor in the ability of Convolutional NNs to outperform traditional methods \citep{guilloteau2020}. However the myriad of filters learned by Convolutional NNs may not always be necessary. Instead, concepts from classic image processing can be used to extract spatial information \citep{hilburnMadison}. For example, key spatial information can be extracted (1) using predefined filters to extract specific image properties, e.g., the local mean or gradient of an image; and (2) image pyramids that represent images at different resolutions \citep{burt1983, adelson1984}, which provides a representation of multi-scale information in the images.  We believe that these classic tools are under-utilized ever since CNNs became popular in our research community. Many ML algorithms could be made much simpler, more robust, and more interpretable, through more extensive use of feature generation using these classic methods, specifically by using image pyramids combined with classic filters to generate stronger features that can be used with simpler machine learning methods, e.g., support vector machines or random forest instead of convolutional NNs.

\subsubsection{\bf Harmonic Analysis (Fourier and Wavelet transforms)}
\label{sec:harmonic_analysis}

Another way to extract and utilize spatial information by mathematical means is to apply harmonic analysis to obtain spectral properties of images, e.g., by transforming images using the spatial version of the Fourier or wavelet transforms. 
The properties in spectral space, e.g., the presence of high magnitude signals of certain (spatial) frequencies, can be efficient features for the occurrence of certain patterns.
More complex examples of using harmonic analysis in combination with machine learning for meteorological imagery include the use of harmonic analysis in neural network to focus on specific spatial scales 
\citep{lagerquist2022can} and
the use of Wavelet neural networks \citep{stock2022trainable}.

(The above work is not to be confused with the recent, impressive work on using Fourier operators in neural networks for weather forecasting \citep{pathak2022fourcastnet}, where Fourier transforms are employed for the sole purpose of speeding up computation of the convolution operations, not for the purposes discussed here.)

\subsubsection{\bf Topological data analysis}
\label{sec:tda}

Topological data analysis (TDA) provides several tools to extract topological properties from meteorological imagery.  For example, the topological concept of {\it persistent homology} focuses on the number of connected regions, and the number of holes therein, for a varying intensity threshold in the image, which in turn allows to distinguish different types of patterns, e.g., to classify the mesoscale organization of clouds \citep{hoef2022primer}. 
Persistent homology and other topological properties are emerging in several environmental science and related applications, including in the context of identifying
atmospheric rivers \citep{muszynski2019topological},
Rossby waves \citep{merritt2021visualizing},
local climate zones \citep{sena2021topological},
activity status of wildfires \citep{kim2019deciphering},
and quantifying the diurnal cycle of tropical cyclones \citep{tymochko2020using}.
Generally, TDA is not used as standalone technique, but as a preprocessing step to extract important features, often to be used along with other physically interpretable features, followed by a simple machine learning algorithm, e.g.\ support vector machines.

\subsection{Image features learned by machine learning (generally least transparent)}

An ML algorithm creates its own features during the training process, e.g., a CNN learns its own convolution filters, rather than using pre-defined convolution filters from classic image processing (Subsection \ref{sec:classic_image_processing_tools}). One might be tempted to try to gain an understanding of a CNN by analyzing the individual filters it has learned.  However, that is a challenging task because CNNs use many layers of filters stacked on top of each other that {\it in combination} extract patterns.  Therefore, investigating a single learned filter in isolation is usually not sufficient to reveal its purpose in the entire network.   
This is why many Explainable AI methods, such as attribution maps \citep{mcgovern2019making,ebert2020evaluation}, seek to develop an understanding of the NN's strategies by focusing on the functionality of the entire neural network, rather than of individual filters \citep{olah2017feature}.  Explainable AI is helpful, and sometimes allows one to identify features that a NN is paying attention to \citep{ebert2020evaluation}, but it is not a magic bullet, and certainly not a replacement for designing ML methods that are a priori interpretable \citep{rudin2019stop}.  Thus, we believe that for the great majority of environmental science applications (there are exceptions!) one should seek to maximize the use of meteorologically and mathematically motivated image features first, then apply a simpler ML method that builds on these features. 
This philosophy might run counter to the traditional view in the AI community to leave all learning to the ML algorithm.  
However, this philosophy is often called for by the special needs of typical environmental science applications, such as (1) significant knowledge of key patterns, (2) small number of available training samples, and (3) use as decision support tool for high stake decisions (e.g., severe weather warnings or climate policy decisions) that require a higher standard of transparency.

\section{Convolutional Neural Networks to Analyze Image Sequences}

\label{sec:NN_simple}
\label{sec:CNN}

Once feature engineering methods have been applied as much as possible to create stronger signals, machine learning methods can be used on top of those features.  In this section we focus on convolutional neural networks (CNNs), which have become a very popular tool for this purpose. 

Given a billion training samples with accurate labels, CNNs can learn extremely complex patterns.  However, in meteorological applications there are often relatively few labeled samples available.  Thus, when we use CNNs to extract spatio-temporal patterns from meteorological image sequences, we need to weigh the complexity of the CNN, in particular the number of parameters to be learned, against the number of available training samples.  
This motivates the search for simple architectures that work for a given task, rather than insisting that any type of spatio-temporal pattern can be found.
As we will see, different CNN architectures  make different assumptions about patterns to be recognized, and achieve different trade-offs regarding pattern flexibility versus model complexity. 

In this section we focus on relatively simple CNNs that use as building blocks layers of type conv2D, conv3D, LSTM, and convLSTM.  The names of these layers are from the TensorFlow/Keras (\cite{tensorflow2015-whitepaper}) 
programming environment. Equivalent layers exist in PyTorch and other standard neural network development frameworks.
We assume that readers are familiar with the basic concepts of CNNs, such as 2D convolution layers, pooling layers, etc.  See \citep{cnn, cnn2} for an introduction to those topics.

This section is organized as follows.  We first introduce 
spatial convolution layers and several tricks for how to use them for image sequences (Section \ref{sec:building_blocks}). 
Section \ref{sec:convLSTM} discusses recurrent neural networks, both LSTM and convLSTM. 
The remaining sections illustrate how these building blocks can be combined into architectures (Section \ref{sec:first-space-then-time}), and compare the practical use of conv2D, conv3D and convLSTM layers for the convection application (Section \ref{sec:comparison}).

\subsection{Spatial convolution layers and tricks to use them for image sequences}
\label{sec:building_blocks}

In this section we discuss spatial convolution layers, conv2D and conv3D.  
We first illustrate their originally intended use, which was for finding patterns in individual images, then discuss tricks for how they can be used nevertheless to extract spatio-temporal patterns in image {\it sequences}.

\subsubsection{Originally Intended Use of conv2D, conv3D}
\label{sec:original_intent}

\begin{figure}
   \centering
   \includegraphics[width = 0.42 \textwidth]{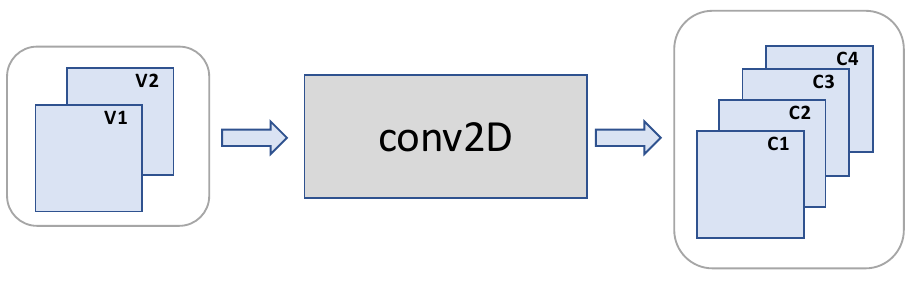}
   
   (a) 2D convolution for planar (2D) image

   \includegraphics[width = 0.5 \textwidth]{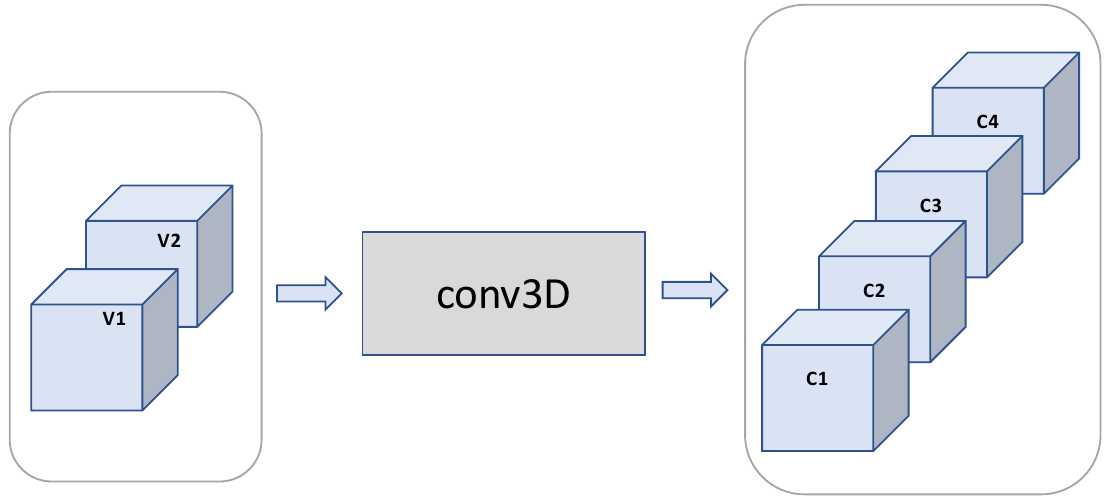}
   
   (b) 3D convolution for spatial (3D) field\\

   \caption{Spatial convolution layers, conv2D and conv3D, were originally designed to extract patterns from an individual image, not from an image sequence. The input is a single image (2D image in (a), 3D image in (b)), and the output is a set of channels, aka activation maps. Each channel corresponds to one spatial pattern, which is learned during training and represented by the filter weights, and tracks the location and strength of occurrence of that pattern in the input image.  The number of input variables (two) and output channels (four) shown here is arbitrary}
    \label{fig:conv_original_use}
\end{figure}

Figure \ref{fig:conv_original_use} illustrates the originally intended use of spatial convolution layers, conv2D and conv3D. As shown in the figure, both were originally designed to extract patterns from individual images.

\subsubsection{Trick 1: Time-to-variable}

While conv2D and conv3D were originally designed to extract patterns from individual images, they have also been used for image sequences in an off-label fashion. The first type of off-label use is by treating time as variable, as illustrated in Figure \ref{fig:time-to-variable}. Here an image sequence is first converted to a single image by treating the images at different time steps as if they were additional variables.
Resulting properties:

\begin{figure}
    \centering

    \includegraphics[width = 0.7 \textwidth]{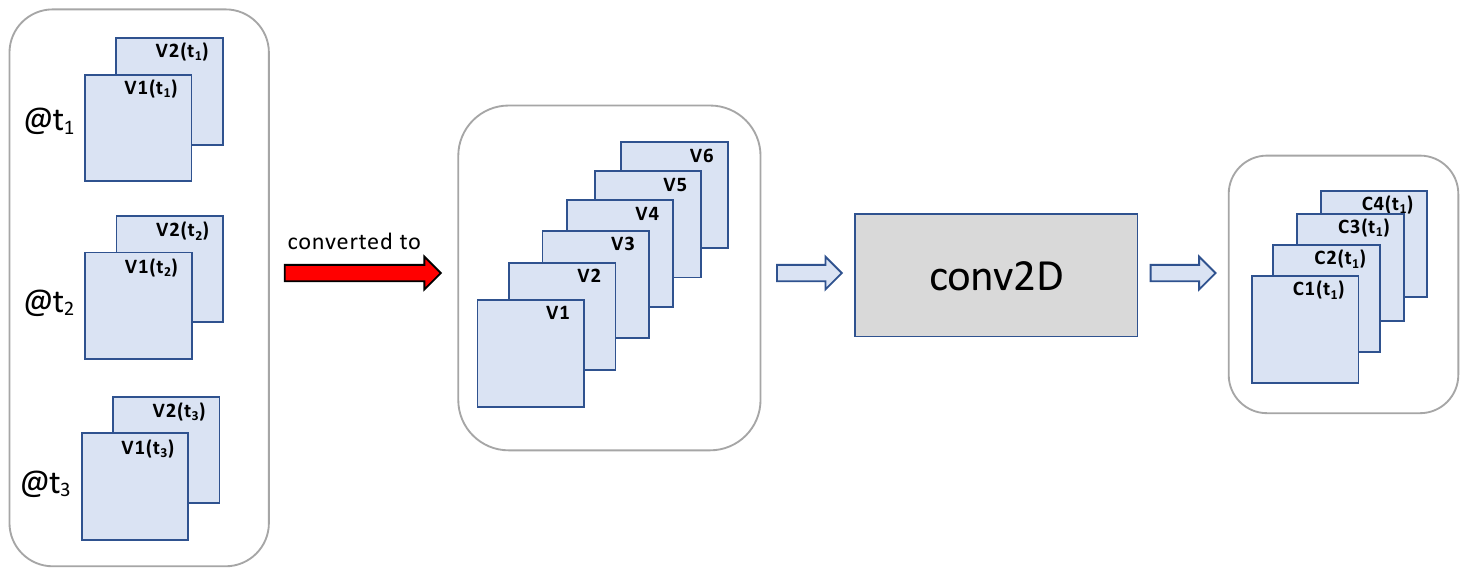}\\
    
    \caption{Time-to-variable: Use of conv2D for a 2D image sequence by first converting the image sequence to a single image with additional variables. The time dimension is thus converted to an increase in variable dimension, i.e.\ time is treated as variables.
    (The same logic can be applied to 3D image sequences by replacing all 2D by 3D images in the schematic, and applying conv3D.)}
    \label{fig:time-to-variable}
\end{figure}

\begin{enumerate}
\item
    The meaning of the order and adjacency of time steps is lost to the network, since they are treated as additional variables, and the order of variables is ignored in neural networks. 
\item 
    At the output of conv2D the time steps are no longer represented separately, i.e.\ the time dimension has collapsed.  Thus all patterns of interest that span different time steps must be identified by this layer.  No other temporal patterns can be identified by a sequential model after conv2D has been applied in this way.
    One should thus carefully consider how early in a network conv2D can be used in this fashion.
\item   
    This approach results in significant dimensionality reduction, so a model with fewer parameters, thus suitable for small sample size.
\end{enumerate}

\subsubsection{Trick 2: Time-to-space}
\label{sec:time-to-space}

In Figure \ref{fig:time-to-space} the time dimension is treated as third spatial dimension. Resulting properties are:

\begin{figure}
    \centering
    
   \includegraphics[width = 0.7 \textwidth]{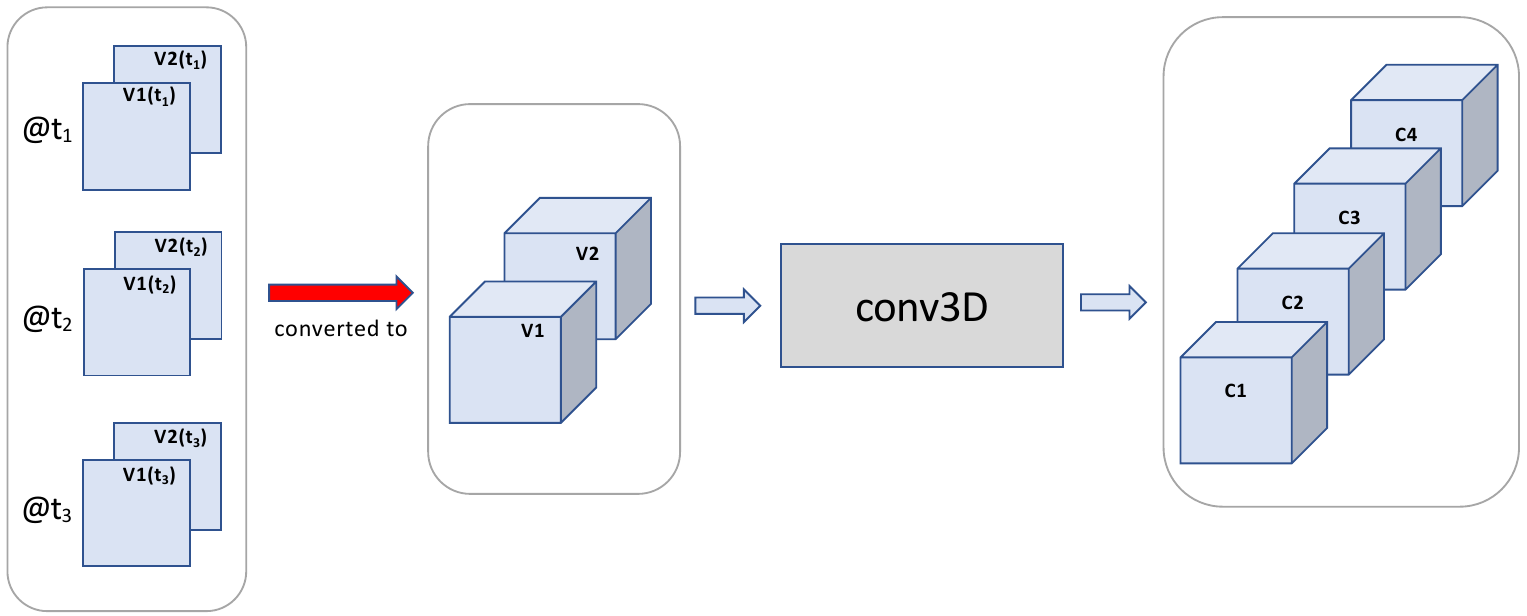}

    \caption{Time-to-space: Use of Conv3D for a 2D image sequence by first converting the time dimension to a third spatial dimension. Although the red arrow seems to suggest that extra conversion is needed to use Conv3D as in Conv2D (Figure 6), this conversion happens \attention{automatically?} 
    in Tensorflow if one uses temporal data with Conv3D, and no explicit conversion is needed. However the figure is drawn this way to emphasize that Tensorflow is not aware that the input data are temporally related}
    \label{fig:time-to-space}
\end{figure}

\begin{enumerate}
\item 
    The third spatial dimension is preserved by conv3D, thus time steps are still represented separately in the output, and their order and adjacency are preserved.
\item
    Spatial convolutions have well known boundary effects.  Namely, the information content at and near the image boundaries tends to decrease with each application of a convolution layer. In large images this effect can often be ignored.  However, given that there are often only fewer than 10 time steps considered, this effect could become significant. Thus, if conv3D is applied repeatedly to a time sequence while time is represented as third spatial dimension, it is possible that the patterns involving the first and last images of an image sequence have less of an impact. Unfortunately, the last image is typically the most recent image which tends to be very important, especially for prediction tasks.
\item   
    Model complexity for the time-to-space approach tends to be significantly higher than for the time-to-variable approach, since the time dimension is preserved and carried through to later layers.
\end{enumerate}

\begin{figure}
    \centering
\includegraphics[width = 0.9 \textwidth]{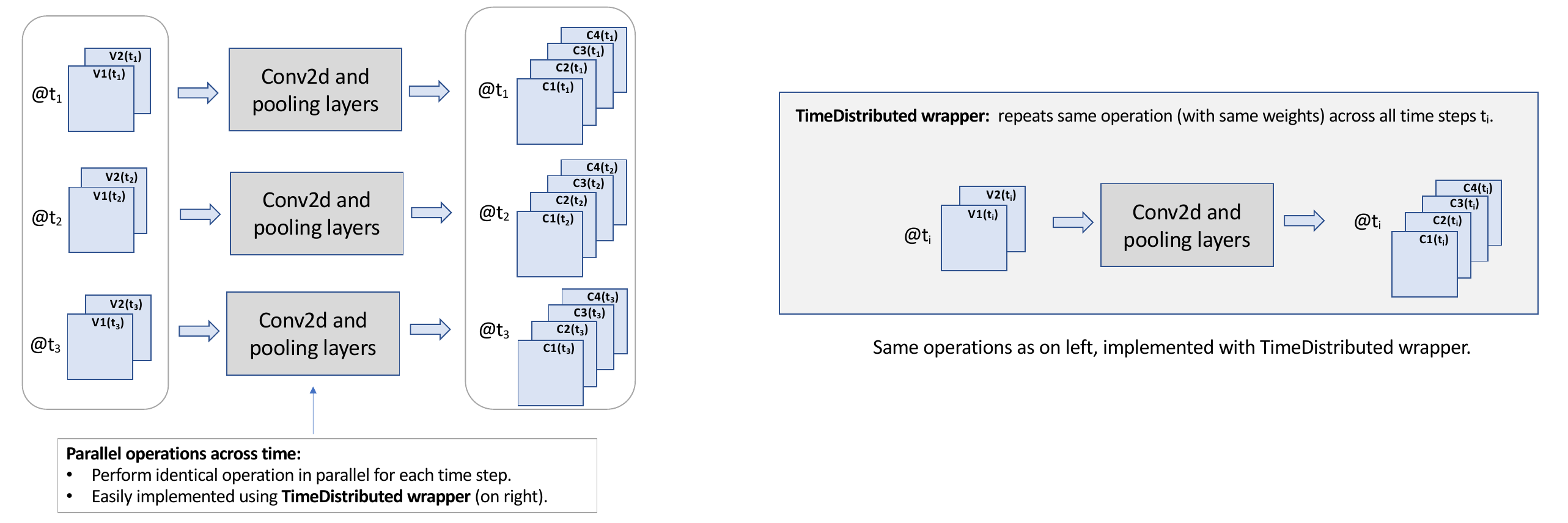}\\

{\small
(a) Details without TimeDistributed wrapper \hspace*{1.0cm} (b) Short version using TimeDistributed wrapper}

    \caption{Use of conv2D in parallel for all time steps using TimeDistributed wrapper.  (a) Expanded view showing identical operations for all time steps; (b) Same operations but using TimeDistributed wrapper to repeat operation}
\label{fig:TimeDistributed}
\end{figure}

\subsubsection{Trick 3: Apply parallel operations for each time step using TimeDistributed wrapper}

\label{sec:time_distributed_wrapper}

Figure \ref{fig:TimeDistributed} illustrates 
one more way of feeding an image sequence into conv2D or conv3D layers, namely using the “TimeDistributed wrapper”. The TimeDistributed wrapper is an easy way to apply the same set of NN layers, such as a convolution layer or any combination of layers, in parallel to the image at each time step.  
The time steps are kept strictly separate, i.e.\ using this method one can extract spatial patterns individually from the image of each time step, but no patterns that involve more than one time step can be extracted.  This operation is thus helpful early in a neural network, especially if one wants to use the first-space-then-time approach outlined in Section \ref{sec:first-space-then-time}.
It should be emphasized that this approach by itself does not extract any spatio-temporal patterns, only individual spatial patterns within each time step, but can be an important element of an architecture that extracts spatio-temporal patterns (see Section \ref{sec:first-space-then-time}).

\subsubsection{Dilated or atrous convolution layer to increase receptive field size}
\label{sec:atrous}

One important concept to consider when designing a CNN model is its {\it receptive field} \citep{araujo2019computing}, i.e.\   
the size of the region in the input image that affects the value of a NN output.
In the context of detecting spatio-temporal patterns the receptive field is important 
because it provides the maximal pattern size in the input image that the NN can "recognize, see   \citep{ebert2020evaluation} for a longer discussion.
One way to increase the size of receptive fields and extract large-scale features is to use a convolution layer with a pooling layer or with a stride bigger than one. A dilated (\cite{dilated}) or atrous (\cite{atrous}) convolution layer is another way of increasing the size of the receptive fields without increasing the number of parameters. This is done through merely having holes between the grid points in the convolution filters. Adding holes can enlarge the field of view of a filter and help extract large-scale features. An additional advantage of using dilated convolution is that the resolution of the output image is kept the same because it does not require a pooling layer. 

{\bf Environmental science examples:}
Many applications using atrous convolution layers are classification problems. \cite{atrous_snow} used atrous convolution layers for classifying snow and cloud, \cite{atrous_cloud} for cloud classification, \cite{atrous_eddy} for ocean eddy detection, and \cite{atrous_land} for land cover classification.

\subsection{Recurrent Neural Networks - LSTM and convLSTM}
\label{sec:convLSTM}
\label{sec:LSTM}
\label{sec:recurrent}

So far we focused on utilizing CNN layer types originally intended to extract spatial patterns, such as conv2D and conv3D.  
In contrast, recurrent neural networks (RNNs) are a special type of neural network developed specifically to extract and utilize patterns in time series, e.g., to be able to predict a value at the next time step given values at prior time steps.
RNNs were first developed for scalar time series (\cite{rnn1,rnn2}). 
Standard RNNs suffer from having a short memory, i.e.\ values from the past are assumed to quickly lose relevance as time progresses.  This can be a problem for many applications, which sparked the creation of the Long-Short Term Memory (LSTM) layer.
LSTM is a recurrent neural network, first introduced by \cite{lstmpaper}. It was developed to overcome the vanishing gradient problem and to carry memory of early time steps through to much later time steps - thus the name {\it long-short term memory}.  
LSTM layers are designed to learn which information to keep or forget, which is achieved by using an internal cell state that passes down relevant information from the previous cells using three gates (forget/input/output gate, see \cite{gates}).  LSTMs have become the dominant type of RNNs used in environmental science applications.

\begin{figure}
\centering
    \includegraphics[width = 0.5 \textwidth]{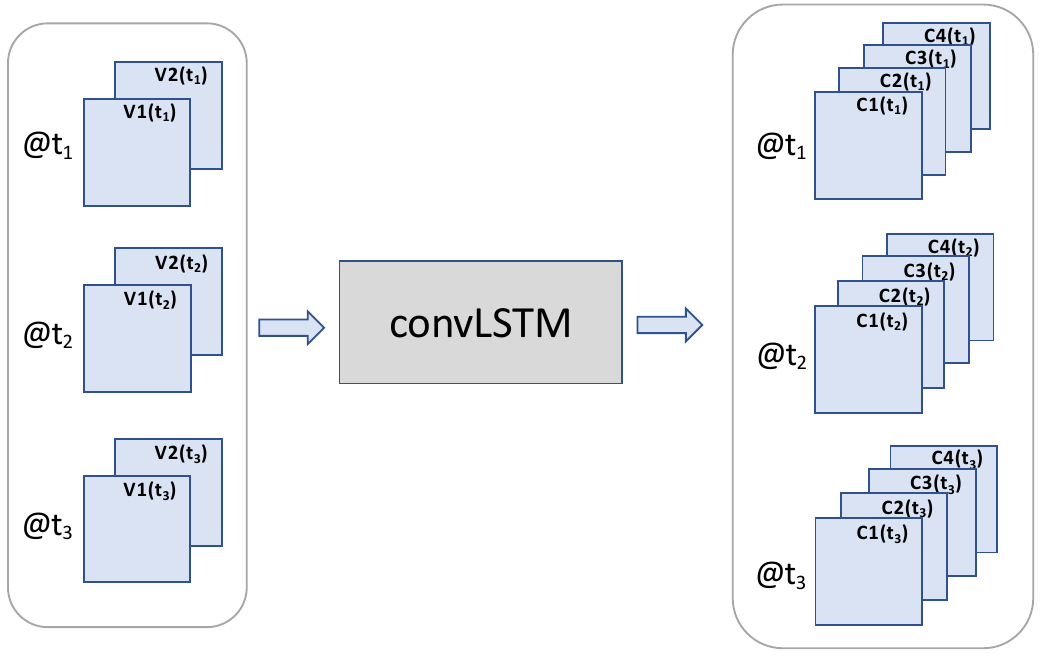}
    
   \caption{The convLSTM layer was designed specifically to extract spatio-temporal patterns from an image sequence, e.g., for predicting the next image in a sequence given prior images. The output of the convLSTM layer shown here has as many time steps as the input. This form is usually used in earlier CNN layers. Another form, which is usually used toward the end of the CNN, outputs a single image, e.g., the predicted image at the next time step.  A hyper parameter, chosen by the CNN developer, determines which form is used for each convLSTM layer}
    \label{fig:convLSTM}
\end{figure}

Since LSTM layers are designed to extract temporal patterns only from {\it scalar} time series, how can they be used for {\it image} time series? 
There are two primary options:
\begin{enumerate}
\item 
    LSTM layers can be used after spatial patterns have been reduced to scalar features (e.g., see Figure \ref{fig:Akansha_original_image}).  This approach is discussed in Section \ref{sec:first-space-then-time}.
\item
    The LSTM layer concept has been expanded to images, leading to convLSTM layers. Namely, the ConvLSTM operator (Figure \ref{fig:convLSTM}) is a convolution version of the LSTM operator developed for video or image sequences to extract spatio-temporal features. This approach is discussed below.
\end{enumerate}

The remainder of this subsection discusses the second option above, convLSTM layers (\cite{xingjian2015convolutional}).
First, for the benefit of those readers familiar with LSTM, but not convLSTM, we briefly describe the difference. 
LSTM is generalized to convLSTM as follows:
In LSTM layers all gates are vectors, i.e.\ 2D tensors including channel dimension. 
In convLSTM layers all gates are images, i.e.\ 3D tensors including channel dimension. Consequently, the matrix multiplications between weight matrices and gates in LSTM are replaced by convolution operations in ConvLSTM. ConvLSTM developed by \cite{xingjian2015convolutional} is already implemented in TensorFlow. Lastly, note that ConvLSTM is based on the LSTM variant with so-called {\it peephole connections} developed by \cite{peephole}, which adds previous cell state information in the gates. 

Resulting properties of using convLSTM layer are:
\begin{enumerate}
\item 
    The time dimension is preserved by the convLSTM layer type shown in Figure\ \ref{fig:convLSTM}, i.e.\ the form that outputs all time steps.
\item
    The number of parameters of convLSTM tends to be higher, comparable to the time-to-space approach in Section \ref{sec:time-to-space}.
\item
    The logic in convLSTM layers is much more complex than in conv2D and conv3D.  Overall, we found it much trickier to train a convLSTM layer successfully than any of the other layers, and a larger number of training samples was needed.
\end{enumerate}

{\bf Environmental science examples:}
ConvLSTM is widely used in meteorological field as the future weather depends on temporal evolution of the previous weather system. Precipitation forecasting (\cite{xingjian2015convolutional,kim2017deeprain,yoonjin3,akbari2019advanced,wang2018application}) and hurricane forecasting (\cite{kim2019deep,udumulla2020predicting}) are perfect examples of its application.

\subsection{An organizing principle: First space, then time}
\label{sec:first-space-then-time}

\label{sec:CNN-LSTM}

\begin{figure}
    \centering
    \includegraphics[width = 0.95 \textwidth]{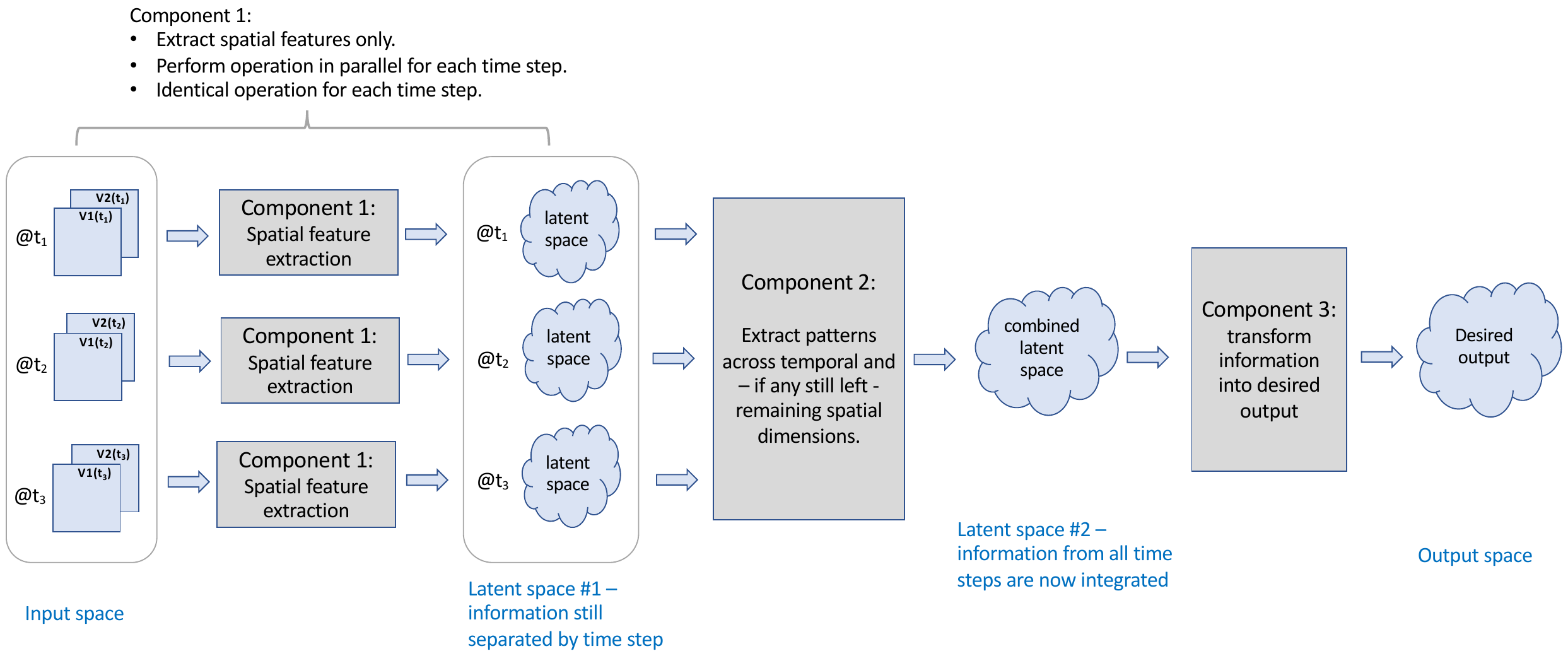}
    \caption{Generic structure of the first-space-then-time approach.  Components can consist of NN layers or other ML methods} 
    \label{fig:first_space_then_time_generic}
\end{figure}

In our search for trade-offs between flexibility in patterns to be recognized versus model complexity, one approach stands out as very common in literature, namely what we call the {\it first-space-then-time} approach.
This approach is based on the observation that images typically contain significant redundant or irrelevant information, i.e.\ not all values of all pixels are important. This tends to be true for meteorological imagery as well.  Thus, one may try to reduce dimensionality in the spatial dimension first, as indicated by the generic architecture in Figure \ref{fig:first_space_then_time_generic}.

The architecture consists of three components: the first component extracts spatial information from the input sequence, resulting in Latent space 1.
(A latent space is an internal NN representation of all the relevant fields 
in a specific NN layer,
which arises 
as the data passes through the subsequent layers in the neural network.) 

The second component extracts the remaining spatio-temporal patterns from Latent space 1, and the third component transforms the resulting spatio-temporal information from Latent space 2 into the desired output. 
The key question to ask for the design of this type of architecture is as follows:

    {\it How much spatial information can we meaningfully extract from each image {\it separately} in Component 1, before taking temporal information into account in Component 2?} 
    
The answer to this question determines how far the dimensionality of the input images can be reduced in the first step, i.e.\ how small we can make the representation of each image in Latent space 1 in Figure\ \ref{fig:first_space_then_time_generic}, before considering time.  
Component 1 is often implemented using the TimeDistributed wrapper discussed in Section \ref{sec:time_distributed_wrapper}.
Component 2, i.e.\ extracting the combined spatio-temporal pattern, is often implemented using i) conv2D with the time-to-variable approach, ii) conv3d with the time-to-space approach, or iii) convLSTM.  
In some cases, namely if the spatial pattern can be expressed in scalars at the output of Component 1, only scalar LSTM layers are needed, rather than a convLSTM layer.
This is the case in the example below.

\begin{figure}
    \centering
    \includegraphics[width = 0.90 \textwidth]{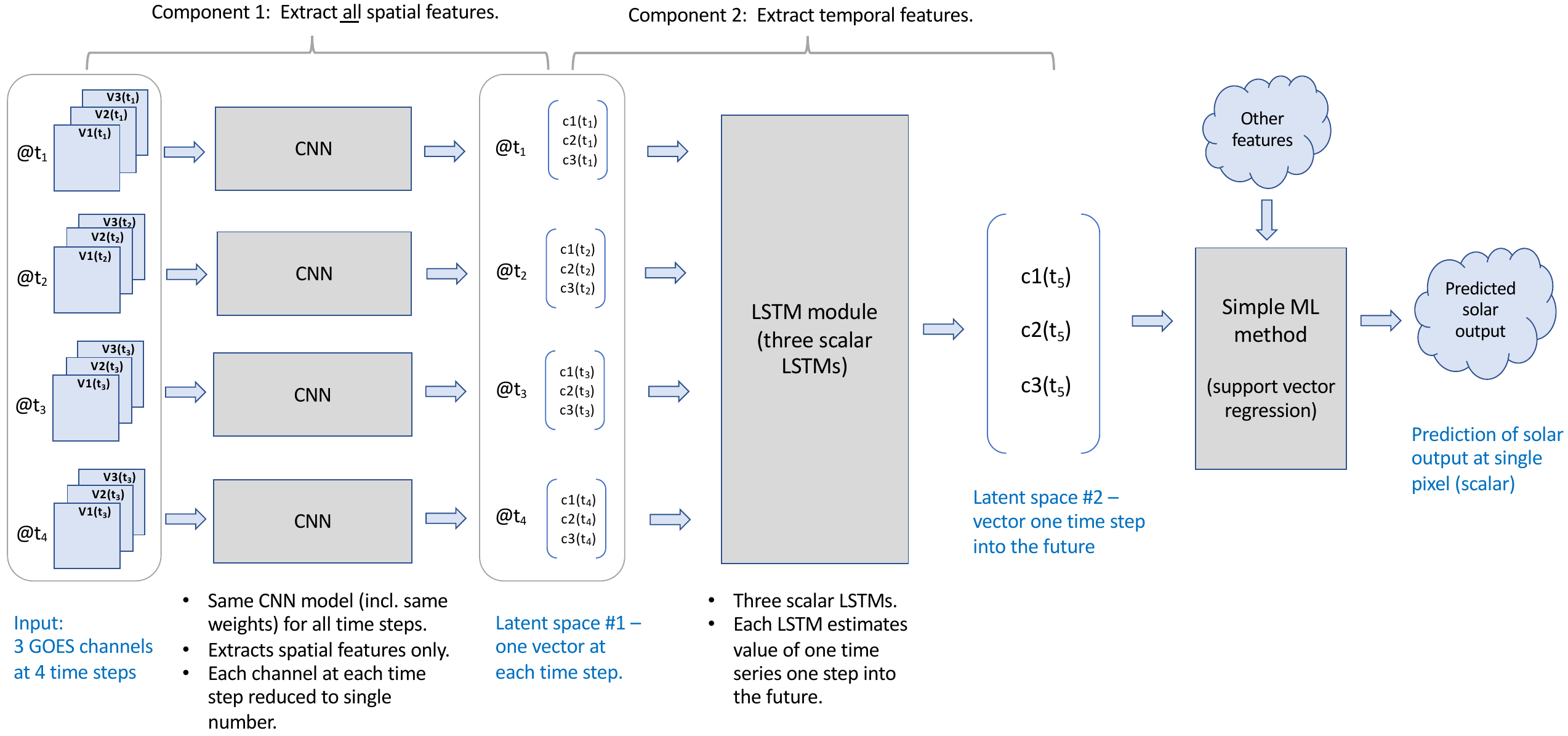}
    \caption{The architecture developed by \cite{bansal2021moment} for solar forecasting uses the first-space-then-time approach.  This is the same architecture as shown in Figure \ref{fig:Akansha_original_image}, but shown in a way that emphasizes its first-space-then-time components}  
    \label{fig:first_space_then_time_Akansha}
\end{figure}

{\bf Environmental science example:}
The solar forecasting application discussed in Section \ref{sec:applications} and introduced in Figure \ref{fig:Akansha_original_image} uses the first-space-then-time principle. 
Figure \ref{fig:first_space_then_time_Akansha} shows that architecture in terms of the  components of Figure \ref{fig:first_space_then_time_generic} from the work by \cite{bansal2021self, bansal2021moment}. The architecture first uses a CNN model to extract the spatial or channel information stored in an image tile and then applies a one-layer LSTM on top of it to extract the temporal information from the time-series visible channels from GOES-16 satellite. Here, the idea is to simply predict the satellite's channel value as a point value in the future. They train CNN spatial extraction and LSTM temporal models to capture the dynamics of the multispectral satellite data. After each image goes for visual feature extraction through the CNN, the per-instant spatial features extracted from CNN over time are passed through LSTM. LSTMs make use of both a cell state, i.e. an internal memory and a hidden state and updates the cell state by combining it with the current input and previous hidden state. By recursively reapplying the same function at every time-step, the LSTM models the evolution of the input features over time. 

\subsection{Comparing the use of conv2D, conv3D, and convLSTM for the convection application} 
\label{sec:comparison}

\begin{figure}
    \centering
    \includegraphics[width = 0.95 \textwidth]{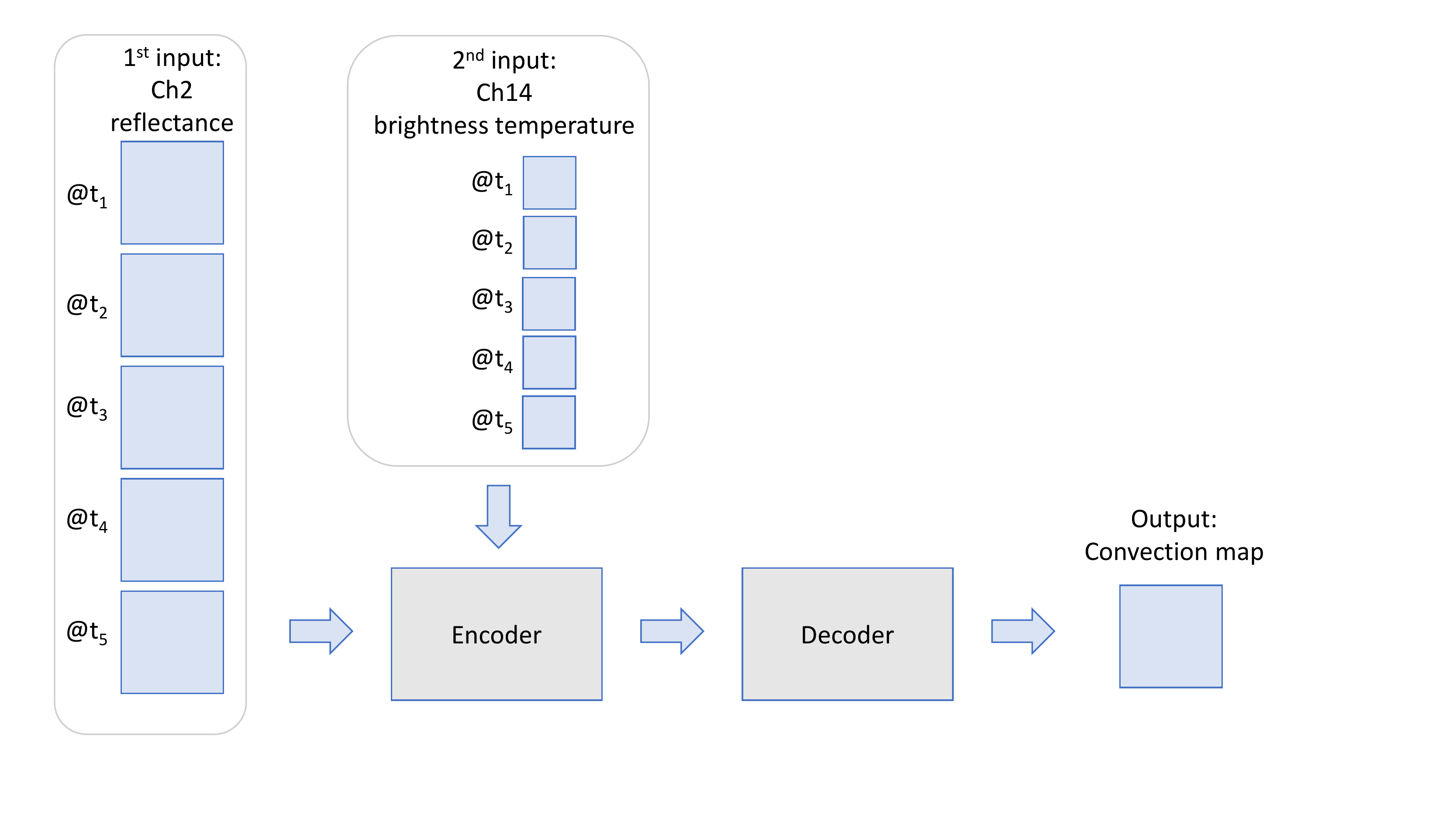}
    \caption{NN architecture used here to compare the use of different convolution layers for a real-world example, namely an encoder-decoder model to detect convection (Section \ref{sec:convection}, \citet{yoonjin4}). 
    A sequence of high resolution input data (Channel 2 reflectance) are ingested in the first layers, and a sequence of lower resolution input data (Channel 14 brightness temperature) are ingested after two maxpooling layers in the encoder part.  We implement the generic encoder and decoder blocks using three different convolution layers, conv2D, conv3D and convLSTM, and compare the results}
    \label{fig:Lee et al. 2021}
\end{figure}

In this section we compare the use of different convolution layers for the example of detecting convection discussed in 
Section \ref{sec:convection} and  by \cite{yoonjin4}. 
Figure \ref{fig:Lee et al. 2021} describes the overall set-up used in \cite{yoonjin4}, which is an encoder-decoder architecture.  Encoder-decoder architectures are common in image-to-image translation problems, where both the input and the output is an image or image sequence.  In this application the input is an image sequence and the output is an image, namely a map indicating the probability of convection for every pixel. 
Both encoder and decoder consist of several blocks of convolution and pooling layers.  The purpose of the encoder is to extract spatio-temporal information, and the purpose of the decoder is to translate that spatio-temporal information into the output image \citep{ebert2020evaluation}.

\def\boxit#1#2{%
    \smash{\color{red}\fboxrule=1pt\relax\fboxsep=2pt\relax%
    \llap{\rlap{\fbox{\phantom{\rule{#1}{#2}}}}~}}\ignorespaces
}

\begin{table}
    \centering
    \caption{Critical Success Index (CSI) for nine experiments using different convolution blocks and different time intervals in input images. Number of parameters and time per epoch for each experiment when using 10 images is also provided for comparisons. The output image contains convective probability, and thus, it requires to choose a threshold to classify a pixel as convection. In this table we list the maximum CSI that results from maximizing over convective probability thresholds. The  number of samples (image sequences) used to train, validate and test,  
    is $10019$, $9192$ and $7914$, respectively, regardless of time interval.
    The red box indicates the result from \cite{yoonjin4}, which used  five images with 2-minute intervals and Conv2D.}
    {\small
    \begin{tabular}{c c c c| c c}
       \hline
       {  } & \makecell{\bf Max CSI \\ \bf 3 images \\ \bf (4min interval)} & \makecell{\bf Max CSI \\ \bf 5 images \\ \bf (2min interval)} & \makecell{\bf Max CSI \\ \bf 10 images \\ \bf (1min interval)} & \makecell{\bf Number of \\ \bf parameters \\ \bf (using \\ \bf 10 images)} & \makecell{\bf Time \\ \bf per epoch \\ \bf (using \\ \bf 10 images)} \\
       \hline \hline
       {\bf Conv2D} & 0.60 & \boxit{0.25in}{.09in} 0.62 & 0.62 & 744,817 & 11s\\
       \hline
       {\bf Conv3D} & 0.62 & 0.65 & 0.67 & 2,205,121 & 186s \\ 
       \hline
       {\bf ConvLSTM2D} & 0.62 & 0.59 & 0.39 & 2,942,321 & 364s \\
       \hline 
    \end{tabular}
    \label{tab:experiment_results}
}
\end{table}

This study greatly extends the study by \cite{yoonjin4}.  Namely, \cite{yoonjin4} only looked at using conv2D layers in the encoder and decoder blocks, while here we also explore the use of conv3D and convLSTM blocks for that purpose.
As shown in Figure \ref{fig:Lee et al. 2021}, each model takes temporal image sequences from a geostationary satellite, GOES-16, to detect convection, and is trained against MRMS data. The input data are a sequence of five images of Channel 2 reflectance, and of five images of Channel 14 brightness temperature, both of which have two-minute intervals. 
As shown in Figure 6, Channel 2 reflectance data are inserted in the beginning, and after two maxpooling layers in the encoder Channel 14 brightness temperature data are inserted, because Channel 14 data have four times coarser spatial resolution than Channel 2 reflectance data. Therefore, in this model setup, five Channel 2 images are treated as five different variables in the input layer, and five Channel 14 images are added as five different channels in the hidden layer. One advantage of this approach is that the training is fast and results are shown to be comparable to the radar product.

For this extended study we designed nine experiments to explore the impact of using different convolution blocks and different time intervals between the input images.  
Namely, we use three different convolution layers (conv2D, conv3D, and convLSTM2D) and 
three different datasets (three images with 4-minute intervals, five images with 2-minute intervals, and ten images with 1-minute intervals). We use the overall encoder-decoder model from \cite{yoonjin4} but need to make adjustments for the convLSTM approach: 
for the conv2D and conv3D there are always two convolution layers before each pooling layers, but when using convLSTM2D layers there is only one convolution layer before each pooling layer due to the large number of parameters and long training time of convLSTM architectures.

Table \ref{tab:experiment_results} summarizes convection detection skills from each experiment in terms of critical success index (CSI; \cite{csiref}), along with the number of parameters and time to train each network for the 10 image dataset. 
Observations include: 
\begin{itemize}
\item 
   {\bf Maximizing CSI:}
   Conv3D performed best throughout,  consistently achieving the highest CSI for all datasets (only comparable to convLSTM for one dataset). 
   Optimal performance was achieved using ten images with Conv3D layers.
   Conv2D still gave decent results, while convLSTM failed miserably for the 10-image input. 
\item
   {\bf Model complexity:}
   conv2D is the clear winner in terms of model complexity: it has the smallest number of parameters (roughly by a factor of three) and training time is more than an order of magnitude shorter than for the alternatives.
   conv3D is an order of complexity higher, but still less complex than convLSTM2D, especially considering that we only used half as many convLSTM layers as conv3D in this experiment.
\end{itemize}

Conv3D appears to be the best solution for this application, but that is certainly not the case for all applications.  The fact that the sample size is relatively small here may have put convLSTM at a disadvantage. 
A key take-away is that even though convLSTM was specifically designed to extract and utilize spatio-temporal patterns, other factors, such as small sample size, may not make it the optimal solution.
Generally speaking, convLSTM is a fairly complex architecture and while it may sometimes yield the best solution, it seems to be more finicky and often failed miserably, as demonstrated for the 10-image dataset here. As a result it also seems to take much more expertise to make it work. 
In contrast, conv3D tends to be much more robust and easy-to-use in our experience. 
conv2D is by far the simplest solution. It might often be less accurate, but it is fast, easy-to-use, and requiring few samples, so sometimes might be the model of choice. 
Given that to date there are no rigorous guidelines for which convolution type to use for which application, we encourage developers for now to consider (and maybe try) all three convolution types, keeping the general tendencies above in mind.

\section{Concept of Attention - A Major Leap for AI Applications in Environmental Science}
\label{sec:attention_new}
\label{sec:attention}

In this section we briefly discuss the novel concept of attention, 
starting with its origin in natural language processing, its use for image generation and interpretation, brief discussion of key NN architectures such as transformers and diffusion models, and its emerging impact on environmental science.  
Explaining how the various attention-based models work in detail is far beyond the scope of this manuscript. Instead we only briefly introduce the concepts, with the goal of providing readers with an understanding of the ground breaking new abilities of attention-based models and their potential impact on environmental science applications.

The attention concept has had a tremendous impact on the abilities of AI systems for image-related tasks, as evidenced by three new AI-driven art tools that have stunned both the computer science community and the public in 2022: 
OpenAI's {\it Dall-E-2} \citep{dalle} and Google's {\it Imagen} \citep{imagen} generate photo-realistic images from a text description provided by a user, similarly Meta's {\it Make-A-Video} generates 5-second video clips from text
\citep{metatexttovideo}.  The abilities of these new tools are astounding and were unimaginable just a couple of years ago.
There have been countless headlines in the press since their release in 2022, including articles entitled "Meet DALL-E, the A.I. That Draws Anything at Your Command" (New York Times; 
\citet{NYT_DALL-E_2022})
and "AI can now create any image in seconds, bringing wonder and danger"
(Washington Post; \citet{WP_DALL-E_2022}), with the latter discussing potential risks that come with making these powerful new tools available to the public. 
The use of the attention concept was one of the primary factors that made this success possible, both to interpret text prompts using attention-based Natural Language Processing models (Section \ref{sec:attention_origin}) and to generate realistic images using attention-based diffusion models (Section \ref{sec:diffusion}). 

Given these powerful new abilities of attention-based NNs for both image generation and interpretation, it is not surprising that they are now taking the environmental science community by storm, leading to 
a first workshop dedicated entirely to this topic (Transformers for Environmental Science; \citet{esst}).

{\bf Environmental Science Examples:}
Attention-based methods have already found application for precipitation mapping \citep{metnet1, metnet2}, generating super-resolution imagery \citep{attention11}, wildfire estimation \citep{attention12}, 
population density estimation \citep{attention13}, damage assesment \citep{attention2}, and land cover estimation \citep{attention9,attention10}.

\subsection{The Origin of the Attention Concept - Natural Language Processing}

\label{sec:attention_origin}

Attention-based neural networks have revolutionalized the field of Natural Language Processing (NLP) over the past 5-7 years, e.g., providing a new generation of algorithms for translating text from one language to another. 
In a translation task a sentence is typically interpreted as a temporal sequence of words.  The challenge is that each word may have multiple meanings in a language, and the correct meaning must be determined from its context, i.e.\ from the other words in the sentence (or even from words in prior sentences), before it can be translated correctly.  
Attention-based NLP algorithms address this by identifying for each word in a sentence 1) which other words in the sentence provide highly relevant context for its meaning (i.e.\ identify a weighting factor for the other words relative to the considered one), and 2) which word meaning each of the other words points to.  The meaning of the considered word is then estimated as a weighted sum of the meaning pointed to by all the other words in the sentence. The weight factors of the other words here define how much {\it attention} is given to each one of them, i.e.\ how much context it provides for the word of interest.
This forms one of the core ideas of the attention concept - a flexible way to assign weighting to elements that are even far away from the currently considered element.  In contrast, in traditional RNNs and CNNs the context for the interpretation tends to be limited to other elements in close temporal or spatial proximity.  
A key question, of course, is how attention is implemented, i.e.\ how attention layers can learn efficiently which other words in the sentence we need to pay how much attention to for the interpretation of a considered word.   
However, space limitations prevent us from providing the details of how attention is implemented - we refer the interested reader instead to \citep{attention6, attention7, attention8}.

\subsection{Use of Attention for Meteorological Imagery}

\label{sec:diffusion}

Following the success in NLP the concept of attention, which led to novel neural network architectures like {\it transformers} \citep{vaswani} and {\it diffusion models} \citep{diffusion}, has resulted in major breakthroughs in image interpretation and image creation. 
One of the earliest and simplest ideas to transfer the NLP approach to image analysis is to subdivide each image into small non-overlapping patches, then treat the spatially-arranged patches within an image analogous to the temporally-arranged words in a sentence.  
This is the idea behind the article entitled "An Image is Worth 16 x 16 words: Transformers for Image Recognition at Scale" by \citet{dosovitskiy2020image}, which has already been cited over 7,000 times since it appeared in 2020.  
Many more complex analogies and extensions have been proposed that enable the use of attention not only for images, but also for image sequences, and not only for interpretation but also for generation of images, as demonstrated by the aforementioned AI art tools, {\it Dall-E-2}, {\it Imagen}, and {\it Make-a-Video}, and discussed for meteorological imagery below.

\subsubsection{Attention for Interpretation of Meteorological Imagery}

Attention-based models have huge potential to better leverage
the multitude of information stored in the spatio-temporal patterns of meteorological imagery. These new architectures can learn more complex spatio-temporal patterns, including recognizing and modeling patterns that extend over large spatial and temporal distances,  {\bf Such models thus have the potential to produce good estimates of 
physical processes over longer time horizons.} This can often be achieved without increasing the complexity of the model in terms of number of parameters to be learned, making this a very attractive solution for applications with limited sample size, but complex patterns. 
In other words, {\bf attention-based models provide the potential to achieve a much better trade-off between pattern flexibility and model complexity than any of the CNNs discussed in Section \ref{sec:CNN}}.
Specifically, CNNs identify patterns that extend over a small distance in both space and time, based on the very nature of convolution filters.  As discussed in Section \ref{sec:atrous}, the spatial extent of patterns can be increased by using multiple convolution and pooling layers, atrous convolutions, 
or adding a fully connected layer, but none of these means are as effective for this purpose as attention layers. 

{\bf The concept of attention frees the NN of the distance limitation in both space and time inherent in regular CNNs and RNNs.} For NLP this ability to access context that is far away is important for sentences, since the context to interpret one word may be words at great distance in the sentence. Likewise, in meteorological imagery important context might be provided by regions that are far away spatially, e.g., due to teleconnections, or far away temporally, e.g., to take into account the temporal evolution of a severe storm.  
In the following we discuss two different means of applying attention for meteorological image interpretation, starting with the simplest approach.

{\bf Adding an attention layer to an existing NN architecture:}
The simplest way to start utilizing the concept of attention is to add an attention layer to an existing architecture. Attention layers now exist in most neural network programming environments and are easy to add to an existing model as an additional layer. 

{\bf Environmental Science Examples:} 

U-Net based attention has been used for damage assessment of buildings after a disaster \citep{attention2} and cloud detection \citep{attention3}. 
U-Nets with bidirectional LSTM and attention mechanism 
\citep{attention10,attention4} leverage features from time-series satellite data to identify temporal patterns of each land cover class and automate land cover classification. 
\cite{attention9} used attention mechanism to learn the correlations among the channels for super-resolution and object detection tasks.

{\bf Using Transformers:}
Transformers are the first models that rely entirely on the concept of attention without using RNNs or convolution layers. These models were first proposed by \cite{vaswani}. (A TensorFlow implementation is available as part of the Tensor2Tensor package \citep{tensor2tensor}.) Transformers focus on learning context from an input stream (temporal sequence) and have become the dominant architecture in the NLP domain. Transformers introduce two new concepts, namely self-attention and multi-headed attention, for details see \cite{vaswani}. 

Transformers applied to images are most commonly called vision transformers (ViT) and the first work to demonstrate ViTs was \citep{dosovitskiy2020image}.

{\bf Environmental Science Examples:} 
The meteorological community has already started leveraging the power of transformers for different applications, including satellite time series classification \citep{transformer1}, 
change detection \citep{transformer2}, landcover classification \citep{transformer4,transformer5,transformer6}, and anomaly detection \citep{transformer7}. 
\citet{EarthformerES} proposes use of transfomers for Earth science application by proposing an efficient space-time transformer for forecasting. 

Transformer based models on imagery can be further categorized as 
time-series transformers \cite{spatiotempotransformer} that are mostly applicable for forecasting, anomaly detection, and classification and work by capturing long range dependencies from continuous stream of meteorological images, or time-series transformers \cite{timetransformer} that work by learning time representations from the data stream or spatial transformers \cite{spatialtransformer} that aim to extract the spatial features from an image. 

\subsubsection{Attention for Generation of Meteorological Imagery}

Generative Adversarial NNs (GANs; \cite{gan}) and Variational Auto-Encoders (VAEs; \cite{vae}) are so-called {\it generative models} that can be used to generate detailed realistic-looking imagery.
Both types have been explored for generating meteorological imagery \citep{rs13214284, Sun_2020}  
Diffusion models \citep{diffusion} are the first attention-based generative models and are outperforming and slowly replacing GANs and VAEs for many (but not all!) image generation tasks.

{\bf Using Diffusion Models:} Diffusion models are trained by adding noise to an input image where the noise mimics a statistical diffusion process. At each step a small amount of noise is added, but the process is repeated several times. The NN model seeks to "undo" the damage done by the iterative diffusion process, by learning how to invert each diffusion process. It is important to understand that  each inverse diffusion process is modeled as a statistical process with parameters learned using a NN approach - thus at execution time diffusion models are slower than standard NN models, as they have to evaluate several statistical processes. The result is a versatile tool with numerous applications in meteorology, e.g., to de-noise imagery, fill in missing data, or convert a low-resolution image into a super-resolution image \citep{diffusion2}. Diffusion models, since implementing statistical models, can provide probabilistic distributions at the output which is advantageous for many types of applications.  The only disadvantage is that execution is resource-intensive and relatively slow, 
which for some real-time applications may be a limiting factor.

{\bf Environmental Science Examples:} 
Diffusion models were originally proposed in 2015 and recently gained lots of interest due to their training stability and high quality results in image and audio generation. These models are robust and work well even in the presence of missing image pixels which is a common chalenge in meteorological data.
In the context of spatio-temporal information rich meteorological data, diffusion models can be particularly interesting for reconstruction accuracy and consistency to train physics-informed neural network that creates physically consistent high-resolution imagery \citep{diffusion1}, solve the problem of super resolution, including improving the performance of downstream models for image classification and segmentation. Diffusion models have shown to be a huge success in demonstrating their ability in generating high-quality image and videos from text but their ability is yet to be explored by the earth science community. However, we expect this to change in the near future.

\section{Other ways to reduce the need for large labeled datasets - Unsupervised, Self-supervised, and Transfer Learning}

\label{sec:training}
\label{sec:training_strategies}
This section focuses on yet another set of strategies that can help us make the most of meteorological image sequences. Most meteorological data tends to be {\it unlabeled}, e.g.\ satellite imagery usually does not come with labels that specify whether or where there is convection, or which land cover types a present. In contrast, {\it labeled} datasets come with such labels. The number of available samples along with the availability of these labels are key factors as to which machine learning model can work best for the target application. Labeled data is generally more sparse and not easy to create, while unlabelled data is often abundantly available. 
In order to fully leverage the more advanced and data hungry models like LSTMs, GANs, transformers, or diffusion models, huge amounts of training samples are needed. This creates a need to utilize different types of ML techniques, including unsupervised, supervised, self-supervised, and transfer learning that can overcome the challenges of data quality and quantity.

{\bf Unsupervised Learning}, aka learning without supervision, trains models on unlabaled datasets. A well known unsupervised method is clustering which groups the unlabeled samples based on similarity relationships between them, i.e.\ the model tries to learn the inherent relationship in the input data without labels or ground-truth. 
The utility of such an approach has been shown by \citep{WANG2019303} in learning crop type mapping without field-level labels, by \citep{denby2020discovering} for cloud organization and \citep{jean2018tile2vec} in learning compressed informative representations from unlabeled remote sensing data that also exceeds the performance of supervised approaches on some tasks.
{\bf Supervised Learning} seeks to learn relationships between each training sample and its label, with the goal of predicting such labels during operation for new samples that come without a label.  The learning is {\it supervised} in the sense that for each training sample the model is given feedback on whether it assigned the correct label, and if not how far it was off. 
{\bf Self-supervised learning} takes advantage of internal relationships in a dataset.  For example, given a temporal sequence of imagery without labels, one can train a model to predict each image, given a certain number of preceding images.  Thus, the method trains against itself as output, thus the name {\it self-supervised}. \cite{bansal2021self, bansal2021moment} designed a self-supervised model to nowcast solar irradiance 15 minutes into the future using a combination of CNN and LSTM models on the visible channels of the GOES satellite. The key idea is that the ML model trains on whole CONUS domain and intrinsically accounts for local geographical features without the need for external supervision or inputs. This pre-trained network was then further applied on solar specific dataset to nowcast site-specific solar power for 15 minutes into the future. Some of other applications of this are shown in \citep{vincenzi2021color, ayush2020geography}
Finally, \textbf{Transfer Learning} has become popular in the meteorological domain to make use of relationships already learned and encoded in ML models trained for similar applications, thus also reducing the need for large labeled datasetes. As the name suggests, this method involves utilizing a model trained on a task similar to the task of interest - preferrably a task where a large labeled dataset is available - and  transferring the extracted relationships to create a new model for the new task. \citep{gooch2020improving} show the use of NN models trained on the famous ImageNet dataset \citep{deng2009imagenet} - a primary benchmark for object recognition research with over three million photos of objects along with object labels - as a starting point for a NN model by copying over the lower layers which are trained to extract basic patterns, e.g., shapes and textures, and only training the last layers of the neural network on meteorological images.  The motivation is the assumption that basic patterns in meteorological imagery form a subset of basic patterns occurring in multi-million photos in ImageNet. \citep{transfer1} showed the use of this technique on environmental application of classifying land use and land cover.

Overall, these strategies have been used extensively to analyze mateorological imagery, but not so much yet for image sequences \citep{jean2018tile2vec, bansal2021moment, bansal2021self, rs11060622}.  We anticipate that these strategies are important and underutilized tools to accelerate work in this area.

\section{Conclusions}
\label{sec:conclusion}
In this paper we discussed various tools to extract spatio-temporal patterns from meteorological image sequences, ranging from feature engineering to attention-based neural networks and transfer learning. In particular, there is significant potential for more progress in this domain with the recent advancements in the computer science domain, including but not limited to, the concept of attention, transformers, and generative models such as diffusion models. Models utilizing these concepts \citep{metnet1, metnet2, deepmind, attentionunet, EarthformerES, mike, fourier} have shown to perform better at capturing longer range spatio-temporal dependencies from meteorological imagery, but have not yet been fully leveraged on image sequences. 
These models show astounding efficiency and skill that could be transformative for many meteorological applications.  However, they are also very new and the jury is still out whether their results have any significant shortcomings compared to traditional physics-based models.

\begin{Backmatter}

\paragraph{Acknowledgments}
We thank Rey Koki (CU Boulder, NOAA GSL) for emphasizing to us the huge potential of atrous convolutions for image applications. 

\paragraph{Funding Statement}
The work by IE and ASB on this project was partially supported by the National Science Foundation under NSF AI institute grant ICER-2019758.  KH and ASB acknowledge support by NOAA grant NA19OAR4320073. 

\paragraph{Competing Interests}
The authors declare no competing interests.

\paragraph{Data Availability Statement}
As a survey paper this article does not contain any new data or code. 

\paragraph{Ethical Standards}
The research meets all ethical guidelines, including adherence to the legal requirements of the study country.

\paragraph{Author Contributions}
Conceptualization: all authors.
Investigation:  ASB and YL.  
Funding acquisition and supervision: KL and IE.
Writing (both original draft and editing): all authors.
All authors approved the final submitted draft.

\paragraph{Supplementary Material}
There is no supplementary material for this article.

\bibliographystyle{apalike}
\bibliography{references.bib}

\end{Backmatter}
\end{document}